\pdfoutput=1

\documentclass[11pt]{article}

\usepackage[final]{acl}

\usepackage{times}
\usepackage{latexsym}
\usepackage{algorithm, algpseudocode}
\usepackage{amsmath} 
\usepackage{tikz}
\usetikzlibrary{arrows.meta,positioning,calc}

\usepackage{booktabs, multirow, array, xcolor, colortbl, pifont}

\definecolor{mygreen}{rgb}{0.27,0.70,0.23}   
\definecolor{myyellow}{rgb}{0.74,0.71,0.02}  

\definecolor{cadmiumgreen}{rgb}{0.0, 0.42, 0.24}

\definecolor{myred}{rgb}{0.7, 0.3, 0.0}
\definecolor{myblue}{rgb}{0.2, 0.3, 0.6}

\usepackage[T1]{fontenc}

\usepackage[utf8]{inputenc}

\usepackage{microtype}

\usepackage{inconsolata}

\usepackage{graphicx}

\usepackage{booktabs, multirow} 

\usepackage{tcolorbox} 

\usepackage{hyperref}

\usepackage{longtable}

\usepackage[dvipsnames]{xcolor}

\title{AgentRouter: A Knowledge-Graph-Guided LLM Router 
for \\ Collaborative Multi-Agent Question Answering}

\author{
\small
    Zheyuan Zhang\textsuperscript{1}\textsuperscript{*}, 
    Kaiwen Shi\textsuperscript{1}\textsuperscript{*}, 
    Zhengqing Yuan\textsuperscript{1},
    Zehong Wang\textsuperscript{1}, 
    Tianyi Ma\textsuperscript{1}, \\
\small
    \textbf{Keerthiram Murugesan \textsuperscript{3},
    Vincent Galassi\textsuperscript{1},
    Chuxu Zhang\textsuperscript{2},
    Yanfang Ye\textsuperscript{1}\textsuperscript{$\dagger$}} \\
\small
    \textsuperscript{1}University of Notre Dame, 
    \textsuperscript{2}University of Connecticut,
    \textsuperscript{3}IBM Research, \\
\small
    \textsuperscript{*}Equal Contribution 
    \textsuperscript{$\dagger$}Corresponding Author
    \\
\small
    \texttt{\{zzhang42, yye7\}@nd.edu},
}

\begin{document}
\maketitle
\begin{abstract}

Large language models (LLMs) and agent-based frameworks have advanced rapidly, enabling diverse applications. Yet, with the proliferation of models and agentic strategies, practitioners face substantial uncertainty in selecting the best configuration for a downstream task. Prior studies show that different agents and backbones exhibit complementary strengths, and that larger models are not always superior, underscoring the need for adaptive routing mechanisms. Existing approaches to agent routing, however, often emphasize cost efficiency while overlooking the fine-grained contextual and relational structure inherent in QA tasks. In this paper, we propose \textbf{\textsc{AgentRouter}}, a framework that formulates multi-agent QA as a knowledge-graph–guided routing problem supervised by empirical performance signals. Specifically, we convert QA instance into a knowledge graph that jointly encodes queries, contextual entities, and agents, and then train a heterogeneous graph neural network (GNN) to propagate information across node types and produce task-aware routing distributions over agents. By leveraging soft supervision and weighted aggregation of agent outputs, \textbf{\textsc{AgentRouter}} learns principled collaboration schemes that capture the complementary strengths of diverse agents. Extensive experiments demonstrate that our framework consistently outperforms single-agent and ensemble baselines, while generalizing across benchmarks and LLM backbones. These results highlight the effectiveness and robustness of graph-supervised multi-agent routing for question answering. Our code repo is available \href{https://anonymous.4open.science/r/AgentRouter}{here}. 

\end{abstract}


\begin{figure*}[!ht]
	\centering
	\includegraphics[width=1\linewidth]{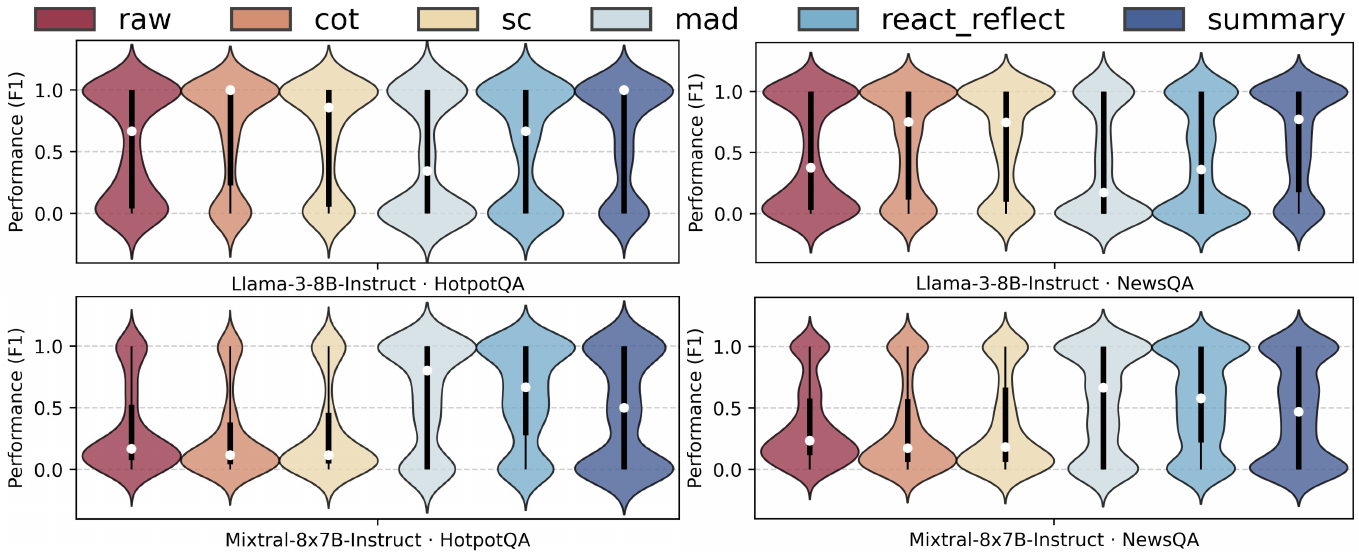}
        \vspace{-20pt}
	\caption{\textbf{Our Motivation.} Performance variance across six classical agent designs (raw, CoT, self-consistency, MAD, ReAct-Reflect, and summary), each applied with \textit{the same} prompt template. Results are shown for two LLM backbones (Llama-3-8B-Instruct, Mixtral-8$\times$7B-Instruct) on two benchmarks (HotpotQA, NewsQA). Each violin plot depicts the distribution of F1 scores across test instances, with the white dot indicating the median and the black bar the interquartile range. \textbf{The plots highlight agents with the same backbone yield wide and non-overlapping distributions, and the relative ranking of agents varies substantially across tasks and backbones.}}
        \vspace{-15pt}
    \label{fig:agent_variance}
\end{figure*}

\section{Introduction}

Large language models (LLMs) have quickly established themselves as versatile learners, demonstrating strong performance in reasoning, comprehension, and natural language generation. Their rapid advancement has significantly influenced both academic research and real-world applications, spanning areas from scientific exploration to software engineering \citep{kojima2022large, ouyang2022training}. Building on this progress, LLM-driven agents have gained prominence for their ability to plan, interact, and solve complex tasks with limited human oversight. These agents extend LLMs into practical pipelines, enabling applications such as program synthesis and repair, retrieval-augmented generation, data analysis, and interactive decision-making \citep{jimenez2024swe, singh2025agentic, guo2024ds, li2024embodied, ma2025autodata}.  

Despite these advances, the rapidly expanding ecosystem of LLMs and agentic strategies introduces a pressing challenge: given a fixed downstream task, practitioners face substantial uncertainty about which model or agent configuration is most effective. Prior studies consistently show that different agents and LLM backbones exhibit task-dependent strengths, and that larger models do not always outperform smaller ones in every scenario \citep{chen2024agentverse, chen2024reconcile}. This heterogeneity underscores that a one-size-fits-all solution is inherently suboptimal, motivating the development of adaptive routing mechanisms.  

\noindent\textbf{Major Research Gaps.} To address this challenge, recent research has explored LLM selection and routing \citep{jiang2023llm, ong2024routellm, chen2024routerdc}. These studies highlight the potential of adaptive selection, yet important limitations remain. \textbf{First,} most prior approaches neglect the rich structural context underlying QA tasks. Many ignore semantic information in the input context altogether, and even works that attempt to incorporate supervised signals (e.g., \citep{feng2024graphrouter}) still fall short in modeling the fine-grained contextual structures that drive effective reasoning. \textbf{Second,} much of the literature emphasizes cost efficiency or adaptability across heterogeneous tasks. While valuable, this perspective overlooks another prevalent scenario: the downstream task is fixed, but new data continuously arrives, requiring the optimal collaboration scheme beyond the single best agent that maximizes future performance.

To mitigate the aforementioned gaps, we propose \textbf{AgentRouter}, a framework that formulates multi-agent Question Answering (QA) as a knowledge-graph based routing problem guided by supervised signals. Our approach proceeds in two stages. First, we construct a knowledge graph in which queries, contextual entities, and agents are jointly represented, with edges encoding lexical, semantic, and relational signals. Prior research has shown that knowledge graphs are particularly beneficial for tasks involving multi-hop reasoning or relational context \citep{jin2024graph, jiang2024kg, peng2024graph}, making them a natural fit for routing in complex QA. Second, we propose a heterogeneous graph neural network (GNN) over this knowledge graph setting to propagate information across node types and to produce task-aware routing distributions over agents. The router leverages soft supervision derived from empirical agent performance, learning distributions rather than hard assignments, and generates final predictions through weighted vote aggregation of agent outputs.  

\noindent\textbf{Why it matters?} By embedding queries, entities, and agents into a unified graph, \textsc{AgentRouter} grounds agent selection in the same semantic structures that govern reasoning for QA. By learning collaboration schemes from supervised graph signals, rather than relying on heuristic voting or LLM-based judges, our method adapts to new inputs while effectively capturing the complementary strengths of diverse agents. Extensive experiments confirm that our approach consistently outperforms single-agent and ensemble baselines, while also generalizing across benchmarks and backbones. These results underscore the effectiveness and robustness of graph-supervised multi-agent routing. Our contributions can be summarized as follows:

\begin{itemize}
    \item\textbf{KG-based Formulation of Multi-Agent QA.} We present the first framework that converts multi-agent question answering into knowledge graphs, where nodes represent not only queries and agent prompts but also fine-grained entities and contextual interactions, thereby preserving and exploiting latent semantic dependencies critical for complex QA.
    \item\textbf{Graph-Supervised Collaboration Learning.} We propose a framework that learns an adaptive collaboration scheme across diverse agent designs and  LLM backbones, which leverages supervised graph signals instead of LLM-based judges or heuristic ensembling.
    \item\textbf{Robust Empirical Validation.} Extensive experiments show that our collaboration scheme consistently outperforms the best single agent and state-of-the-arts baselines.
\end{itemize}

\section{Problem Formulation}

\noindent\textbf{Overarching Goal. }We study the problem of designing an optimized LLM-based agent router for a fixed downstream task. Let $\mathcal{A}=\{a_1,\ldots,a_n\}$ denote the pool of available agents (each agent defined by a backbone LLM and an interaction strategy/prompting style), $\mathcal{X}$ the input/query space, and $\mathcal{Y}$ the output space. For $x\in\mathcal{X}$, each agent $a\in\mathcal{A}$ produces a candidate $y_a(x)\in\mathcal{Y}$. Our goal is to learn an optimal weighted combination of agents in $\mathcal{A}$ that maximizes task-level performance in the fixed downstream setting.  

\noindent\textbf{Assumption and Empirical Validation. }Our formulation builds on the assumption that no single agent or backbone uniformly dominates; rather, their strengths and weaknesses are task- and backbone-dependent. This premise is supported by extensive prior work, which consistently shows that different LLMs or prompting strategies excel in different scenarios. Such heterogeneity has motivated collaborative frameworks, where combining diverse models yields higher overall performance than any single constituent \citep{chen2024agentverse, chen2024reconcile}. We further validate this assumption through our experiments (Figure~\ref{fig:agent_variance}). Specifically, we observe that for a fixed backbone and dataset, agent F1 distributions are wide with non-overlapping confidence intervals, and the agent that performs best on one benchmark is often suboptimal on another. Even under identical prompt templates, performance rankings vary substantially across benchmarks and backbones. These findings substantiate the need for a principled, task-aware routing mechanism rather than a one-size-fits-all collaboration scheme.

\noindent\textbf{Contextual Modeling via Knowledge Graphs. }To capture the contextual information required for routing, we represent the interaction space as a knowledge Graph (KG). Formally,
\[
\mathcal{G}=(\mathcal{V},\mathcal{E}), \quad \mathcal{V}=\mathcal{V}_Q \cup \mathcal{V}_A \cup \mathcal{V}_E,
\]
where $\mathcal{V}_Q$ denotes query nodes, $\mathcal{V}_A$ agent nodes, and $\mathcal{V}_E$ entity nodes capturing contextual information. Edges encode different types of relationships: query–entity and entity–entity edges capture semantic or entity-level relations derived from the input, while query–agent edges capture agent responses or performance signals. The statistics of the KGs across benchmarks can be seen in Table~\ref{tab:benchmark_stats}. Within this KG, the routing problem reduces to learning a function $f_\theta$ that scores query–agent pairs by propagating signals along graph edges:
\[
s(q,a) = f_\theta(q,a;\mathcal{G}),
\]
where $s(q,a)$ estimates the utility of including agent $a$ when solving query $q$ for the given task. The router then computes a weighted combination of agents:
\[
\hat{y}(q) = \phi\!\left(\{\,y_a(q),\,w_a(q)\,:\, a \in \mathcal{A}\}\right),
\]
with weights $w_a(q) \propto \exp(s(q,a))$, and $\phi$ denoting an aggregation rule such as voting, reranking, or learned fusion. Framing the problem in this way refines the high-level goal of "finding the best collaboration scheme" into the concrete task of learning graph-supervised scores for query–agent pairs, which in turn yield optimized weightings of agents for a fixed downstream task.

\begin{figure*}[t]
	\centering
	\includegraphics[width=1\linewidth]{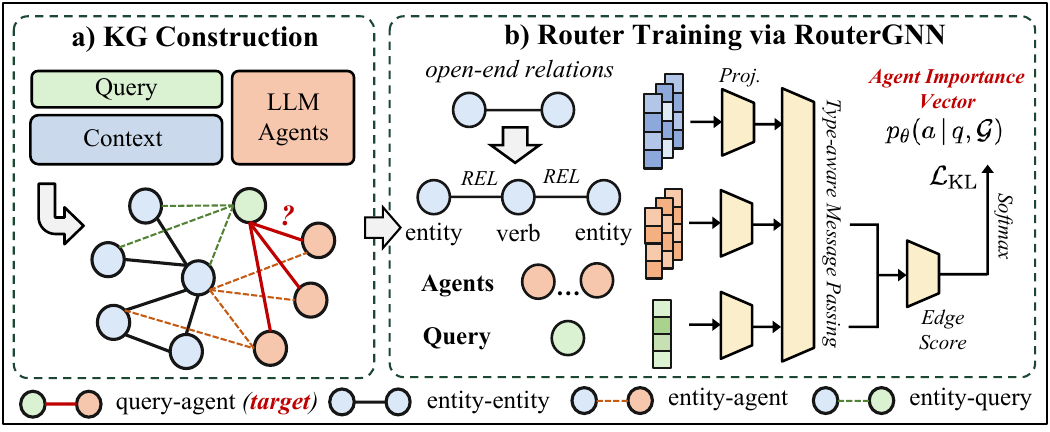}
        \vspace{-20pt}
	\caption{Overview of our proposed framework. (a) QA instances are converted into knowledge graphs with query, entity, and agent nodes, with edges defined to capture semantic relations. Query–agent edges are trainable to enable adaptive routing.  (b) A type-aware heterogeneous RouterGNN propagates contextual and relational information across the graph. The router then predicts a task-dependent distribution over agents, trained via KL divergence against empirical agent performance. Final answers are obtained as the weight vector of the agents per query.}
        \vspace{-15pt}
    \label{fig:framework}
\end{figure*}

\begin{table}[t]
\centering
\begin{tabular}{lcc}
\toprule
\textbf{Statistics} & \textbf{HotpotQA} & \textbf{NewsQA} \\
\midrule\midrule
\multicolumn{3}{l}{\textbf{Avg. \# of Nodes:}} \\
\quad Query           & 1     & 1     \\
\quad Agent           & 24    & 24    \\
\quad Entity          & 111.6 & 46.0 \\
\midrule\midrule
\multicolumn{3}{l}{\textbf{Avg. \# of Edges:}} \\
\quad Entity-Entity            & 331.6 & 110.34  \\
\quad Agent-Entity        & 74.64  & 89.18  \\
\quad Query-Entity    & 22.96  & 3.80   \\
\bottomrule
\end{tabular}
\vspace{-5pt}
\caption{Comparison of knowledge graph statistics between multi-hop and direct QA benchmarks. Agent and query node number are fixed across benchmarks.}
\vspace{-15pt}
\label{tab:benchmark_stats}
\end{table}

\section{Methodology}

Now that we have properly formalized the problem, we proceed to detail our proposed framework. As illustrated in Figure~\ref{fig:framework}, we first transform original query–context pairs into knowledge graphs. These graphs serve as a semantically grounded substrate that captures both contextual evidence and agent-specific perspectives. On top of this representation, we introduce \textsc{RouterGNN}, a type-aware heterogeneous GNN trained to infer query–agent compatibility distributions from empirical performance signals. This design enables the router to move beyond static ensembling, learning adaptive collaboration strategies that exploit complementary agent strengths. 

\subsection{Knowledge Graph Construction}

To enable context-aware routing, we convert each QA instance $(q, C)$, with query $q$ and context $C$, into a knowledge graph $\mathcal{G}=(\mathcal{V},\mathcal{E})$. The central motivation is that the router must reason not only about the query and the available agents, but also about the contextual entities and their relations, since these often determine which reasoning strategies are most effective. Importantly, such contextual information does not directly reveal the best agent; rather, it shapes the structural cues that guide the router in weighting complementary agents. To this end, the graph is designed to contain query nodes, entity nodes, and agent nodes, connected by edges that reflect lexical, semantic, and trainable routing relations.

\noindent\textbf{Nodes as Representations of Context. }  
To faithfully represent the fine-grained contextual information, we design the following node types. 1) Entity nodes are introduced to preserve the contextual signals. We employ a spaCy pipeline to extract named entities, temporal expressions, and numerical mentions from the context, normalizing them into the set $\mathcal{V}_E$. Each entity node stores its surface form, NER type, and frequency, ensuring that repeated mentions amplify the importance of the corresponding node during message passing. 2) Agent nodes form the set $\mathcal{V}_A$, where each node corresponds to one candidate agent defined by its backbone LLM and prompting strategy (e.g., Chain-of-Thought, or ReAct-style reasoning). Representing agents as nodes embeds them into the same latent space as queries and entities, enabling the router to directly learn compatibility signals. 3) Finally, each input question is represented as a query node $v_q \in \mathcal{V}_Q$, whose embedding is initialized via a contextual encoder. This ensures that semantically similar questions occupy nearby regions in the latent space, providing a natural anchor point for supervision.

\noindent\textbf{Edges as Carriers of Relation Signals.}   
Accurate and diverse edges bind the node families to capture both static semantic structure and dynamic routing preferences. Specifically, we design the following types of edges. 1) Query–entity edges link queries to entities they explicitly mention, grounding the question in its evidence context. 2) Entity–entity edges are established through dependency parsing, where we extract relation triples $(h,r,t)$ from spaCy’s parser and record them as connections between entity nodes. This step introduces relational structure into the graph, allowing message passing to propagate semantic information across linked entities. 3) To reflect agent perspectives on context, we further construct agent–entity edges by prompting each agent to identify the most relevant entities it attends to. These edges specify which parts of the context are emphasized by different agents. 4) By contrast, query–agent edges are left as trainable connections. Their role is to carry the routing signal that the model learns to predict: the existence and weight of such edges determine which agents participate in collaboration and how strongly their outputs are weighted. This way, frozen edges capture contextual grounding, and trainable query–agent edges encode adaptive routing.

\noindent\textbf{Edge-to-node Mapping for KG Learning.}  
Since linguistic relations (verbs, dependency types) are open-ended, using them directly as edge types would explode the schema with too many edge categories, but collapsing them into overly coarse classes would also lose valuable contextual information. To solve this problem, we materialize relations as dedicated nodes $\mathcal{V}_{\text{REL}} \subseteq \mathcal{V}_E$. For each extracted triple $(h,r,t)$, we introduce a relation node $r$ and rewire the connection as $h \xrightarrow{\texttt{inc:src}} r \xrightarrow{\texttt{inc:tgt}} t$. This design allows message passing to explicitly traverse relation semantics, while keeping the edge vocabulary manageable. The original text triples are cached for LLM prompting, ensuring that no semantic detail is lost.

\noindent\textbf{Unified Embedding Space and Graph Definition.}  
As such, all nodes are embedded into the shared textual space to make message passing meaningful. Query nodes use contextual embeddings of the question text; Entity and relation nodes use surface text embeddings augmented with type and frequency features; Agent nodes rely on sentence embeddings of their descriptive strategies and prompts. Embedding everything into one space ensures that similarity across node types is preserved before graph-level learning. Although each QA instance yields a unique graph, the agent set is always fixed, thus query–agent edges are consistently trainable. This design ensures that message passing remains effective: structural signals flow through contextual nodes, while routing signals are learned on top of the stable agent-query scaffold. Consequently, we construct the knowledge graph, which provides a principled, semantically grounded representation of QA instances, enabling the router to learn effective collaboration schemes across agents in the next stages.

\begin{table*}[t]
\centering
\resizebox{\textwidth}{!}{
\begin{tabular}{l *{8}{c}}
\toprule
\multirow{2}{*}{\textbf{Method}} & 
\multicolumn{2}{c}{\textbf{2Wiki}} & 
\multicolumn{2}{c}{\textbf{HotpotQA}} & 
\multicolumn{2}{c}{\textbf{NewsQA}} & 
\multicolumn{2}{c}{\textbf{TriviaQA}} \\
\cmidrule(lr){2-3} \cmidrule(lr){4-5} \cmidrule(lr){6-7} \cmidrule(lr){8-9}
& F1 & EM & F1 & EM & F1 & EM & F1 & EM \\
\midrule\midrule
Average        & 51.23$\pm$1.46 & 41.87$\pm$0.91 & 59.52$\pm$1.33 & 46.63$\pm$0.97 & 57.96$\pm$2.01 & 36.11$\pm$0.84 & 45.09$\pm$2.35 & 36.62$\pm$1.05 \\
Majority Vote  & 70.12$\pm$1.95 & 61.00$\pm$1.00 & 63.27$\pm$0.28 & 52.80$\pm$1.12 & 59.11$\pm$0.62 & 39.33$\pm$1.53 & 50.16$\pm$1.19 & 39.00$\pm$1.00 \\
Best LLM       & 71.27$\pm$1.71 & 62.19$\pm$0.94 & 68.32$\pm$1.65 & 50.83$\pm$1.00 & 60.42$\pm$1.74 & 40.79$\pm$0.63 & 48.21$\pm$1.62 & 39.12$\pm$0.44 \\
Best Agent     & 74.89$\pm$1.12 & 63.75$\pm$1.24 & 68.68$\pm$1.70 & 56.20$\pm$1.06 & 62.89$\pm$2.62 & 37.14$\pm$1.41 & 59.33$\pm$2.05 & 49.17$\pm$0.73 \\
\midrule
LLM-Blender    & 66.64$\pm$2.24 & 52.00$\pm$2.65 & 64.97$\pm$1.98 & 50.33$\pm$3.06 & 58.83$\pm$0.66 & 38.00$\pm$1.73 & 48.06$\pm$0.93 & 34.00$\pm$1.73 \\
HybridLLM      & 69.01$\pm$0.15 & 56.00$\pm$1.00 & 59.10$\pm$2.16 & 39.33$\pm$2.89 & 51.96$\pm$2.73 & 28.00$\pm$2.16 & 52.08$\pm$1.36 & 38.67$\pm$1.53 \\
GraphRouter    & 64.09$\pm$0.13 & 56.33$\pm$0.58 & 61.73$\pm$0.76 & 49.67$\pm$0.58 & 64.95$\pm$0.33 & 50.67$\pm$0.58 & 52.37$\pm$1.26 & 42.33$\pm$1.53 \\
\midrule
AgentRouter    & 74.86$\pm$2.34 & 67.33$\pm$0.54 & \textbf{70.54$\pm$0.23} & \textbf{57.33$\pm$0.58} & 65.61$\pm$1.35 & 48.67$\pm$1.53 & \textbf{63.36$\pm$0.19} & \textbf{51.00$\pm$0.88} \\
\quad\quad K=3 & 75.39$\pm$0.40 & 68.33$\pm$0.58 & 68.41$\pm$1.03 & 55.33$\pm$1.53 & \textbf{68.22$\pm$0.84} & \textbf{52.33$\pm$0.58} & 59.91$\pm$1.57 & 49.67$\pm$3.20 \\
\quad\quad K=5 & \textbf{77.02$\pm$0.58} & \textbf{70.67$\pm$0.57} & 69.19$\pm$0.56 & 56.33$\pm$0.58 & \underline{68.13$\pm$0.61} & \underline{51.67$\pm$0.58} & 59.07$\pm$0.71 & 47.67$\pm$1.03 \\
\quad\quad K=10& \underline{76.35$\pm$1.15} & \underline{70.67$\pm$1.15} & \underline{70.07$\pm$0.58} & \underline{56.67$\pm$0.58} & 66.53$\pm$0.93 & 50.00$\pm$1.00 & \underline{62.67$\pm$4.51} & \underline{50.67$\pm$5.86} \\
\midrule\midrule
Oracle         & 92.95$\pm$1.35 & 87.00$\pm$1.73 & 90.70$\pm$1.52 & 83.67$\pm$1.53 & 81.37$\pm$0.17 & 61.33$\pm$1.15 & 71.78$\pm$0.54 & 60.33$\pm$1.53 \\
\bottomrule
\end{tabular}
}
\vspace{-5pt}
\caption{Performance results with baseline methods on the four benchmarks. We report the mean and standard deviation for all results. Best (excluding Oracle) results are in \textbf{bold}, second best are \underline{underlined}.}
\label{tab:main_results}
\vspace{-15pt}
\end{table*}

\subsection{Router Training via RouterGNN}

With the knowledge graph in place, the router must be trained to determine which agents are most useful for a given downstream task. Unlike prior routers that rely on LLMs-as-judges or heuristic voting rules, our training-based approach allows the router to generalize beyond fixed protocols and adaptively weight agents according to contextual signals. This shift from rule-based selection to supervised learning enables the model to exploit nuanced dependencies between queries, entities, and agents that are not accessible through static ensembling.

To achieve this, we adopt \textbf{RouterGNN}, a heterogeneous Graph Neural Network (GNN) that performs type-aware message passing across the knowledge graph. Each node embedding is first projected into a shared latent space with a type-specific projection operator $\mathrm{Proj}_{\tau(v)}$. For an edge $(u \xrightarrow{\psi} v)$ of type $\psi$, the message is computed as
\[
m^{(l,\psi)}_{u \to v} = \mathrm{Proj.}\!\left( W^{(l)}_{\psi}\, h^{(l-1)}_u \right),
\]
and aggregated by mean pooling over all neighbors of type $\psi$. Messages from different edge types are then combined with learnable gates, producing the node update
\[
h^{(l)}_v = U^{(l)}_{\tau(v)}\!\Big(h^{(l-1)}_v \,\|\, \sum_{\psi \in \Psi(v)} w^{(l)}_\psi \cdot \tilde{m}^{(l,\psi)}_v \Big),
\]
where $\tau(v)$ is the node type, $w^{(l)}_\psi$ is a learned scalar per edge type, $U^{(l)}_{\tau(v)}$ is a type-specific update function, and $\|$ denotes concatenation. This update rule ensures that each node embedding integrates both its previous state and type-aware signals from heterogeneous neighbors.

After $L$ layers, the query embedding $h^{(L)}_q$ encodes contextual and relational evidence, while each agent embedding $h^{(L)}_a$ captures its suitability for the query. Routing scores are then produced by
\[
s(q,a) = \mathrm{MLP}\!\left( h^{(L)}_q \,\|\, h^{(L)}_a \right), 
\]
\[
p_\theta(a \mid q,\mathcal{G}) = \mathrm{softmax}_{a \in \mathcal{A}} \big(s(q,a)\big).
\]

Supervision is derived from the empirical performance of candidate agents. For each query, we evaluate all agents and transform their F1 scores into a soft target distribution $p^*(a \mid q)$ via a temperature-scaled softmax, yielding smoother and more informative labels than hard one-hot assignments. The router is then trained by minimizing the Kullback–Leibler divergence
\[
\mathcal{L}_{\mathrm{KL}}(q) = \sum_{a \in \mathcal{A}} p^*(a \mid q)\, \log \frac{p^*(a \mid q)}{p_\theta(a \mid q,\mathcal{G})}.
\]

KL divergence is especially appropriate in this setting. Unlike cross-entropy, which strongly emphasizes only the top-performing label, KL enforces alignment across the entire distribution, ensuring that the router captures the relative strengths of both primary and secondary agents. Compared to mean squared error, KL better respects the geometry of probability distributions and avoids vanishing gradients for low-probability agents. This richer supervision encourages the router not only to identify the best agent but also to approximate the correct balance among complementary agents, leading to more stable optimization and more flexible collaboration policies.

At test time, the router produces a distribution $p_\theta(a \!\mid \! q,\mathcal{G})$ over agents for each query. Final predictions are generated through weighted voting:
\[
\hat{y}(q) = \phi\!\left(\{\, y_a(q),\, p_\theta(a \mid q,\mathcal{G}) : a \in \mathcal{A}\}\right),
\]
where $y_a(q)$ is the output of agent $a$ on query $q$, and $\phi$ denotes a fusion rule such as weighted majority voting. In this way, the router’s learned distribution directly governs how much influence each agent has in the final answer, producing a principled and context-aware collaboration scheme.

\begin{figure*}[t]
	\centering
	\includegraphics[width=1\linewidth]{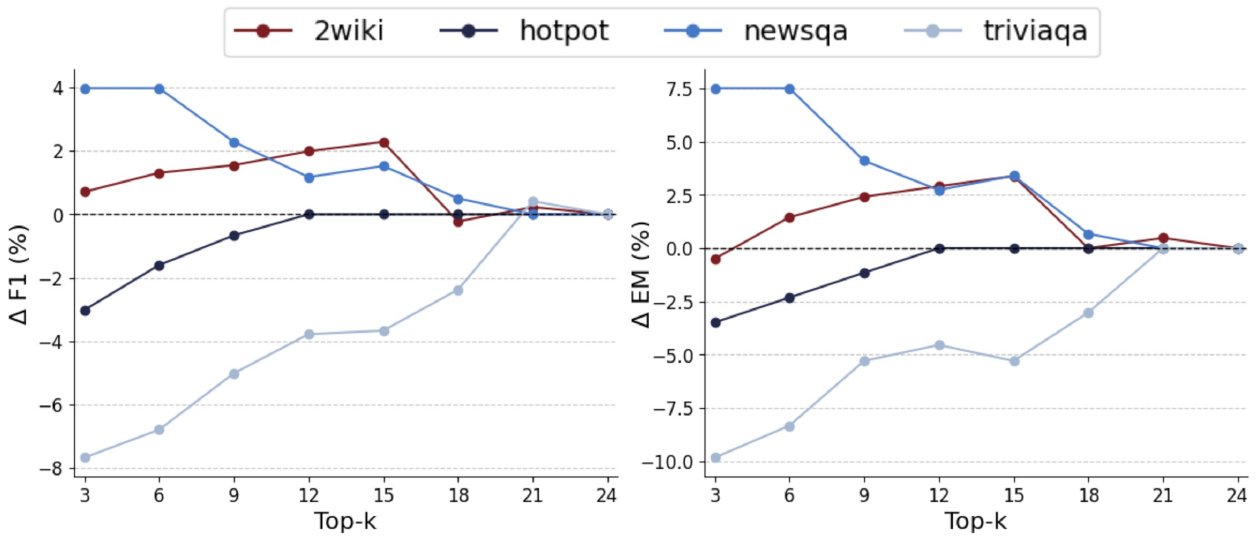}
        \vspace{-20pt}
	\caption{Percentage change ($\Delta$) in F1 (left) and EM (right) relative to $k=24$, used as the base (0\%). Curves show how performance varies with the top $k$ agent clipped across datasets.}
        \vspace{-15pt}
    \label{fig:topk}
\end{figure*}

\section{Experiments}
\subsection{Experiment Setup}
\noindent \textbf{Benchmarks. }We conduct experiments on an extensive collection of Question Answering tasks: HotpotQA \cite{yang2018hotpotqa} and 2WikiMultihopQA \cite{ho20202wikimultihopqa} for multi-hop complex reasoning, and NewsQA \cite{trischler2017newsqa} and TriviaQA \cite{joshi2017triviaqa} for factual and direct reasoning. We describe the details of these benchmarks and discuss the split of sets with other details in Appendix-\ref{appendix:Benchmarks}.

\noindent \textbf{Baselines. } We consider the following baselines: 1) Simple heuristic ensembling methods, including the best LLM method, average score performance, majority vote method, and the best Agent (always choose the agent among the 24 candidates with the best performance). 2) Classic state-of-the-arts LLM routing baselines, including LLM-Blender \cite{jiang2023llm}, HybridLLM \cite{ding2024hybridllm}, and GraphRouter \cite{feng2024graphrouter}. 3) Oracle, indicating the theoretical upper bound, where questions are answered by the best agent. 

While there are many different designs of agents, we hope to cover as many classic and basic design choices as we can. For our experiment purpose, we choose these six agent designs: Raw (the basic LLM method), Chain-of-Thought (CoT) \cite{wei2022chain}, Self-Consistency (SC) \cite{wang2023selfconsistency}, React-Reflection \cite{yao2023react, shinn2023reflexion}, Multi-Agent Debate (MAD) \cite{du2024mad} and Multi-Agent Summary. As demonstrated in Figure~\ref{fig:agent_variance}, different agents perform differently on different LLM backbones. Therefore, we choose four classic LLM backbones of similar scales but from different providers: Llama-3-8b-instruct \cite{meta2024llama3}; Qwen2.5-7B-Instruct-Turbo \cite{qwen2025qwen2_5_7binstruct}; Mixtral-8x7B-Instruct-v0.1 \cite{mistral2023mixtral_8x7binstruct}; and gpt-oss-20b \cite{gptoss2025gpt_oss_20b}. All baseline experiments are run on the same settings and the results are recorded from three consecutive runs to avoid fluctuation. More details of the baselines are in Appendix-\ref{appendix:Baselines}. 

\noindent \textbf{Evaluation Metrics. }Following standard practice on SQuAD \cite{rajpurkar2016squad}, we report Exact Match (EM) and F1. EM measures the percentage of predictions that exactly match the gold answer string after normalization (e.g., lowercasing, punctuation). F1 is the token-level harmonic mean of precision and recall between the predicted and gold spans, capturing partial overlap. EM emphasizes strict correctness, while F1 provides a softer measure that rewards partially correct answers.

\subsection{Main Results}

We present the main results of AgentRouter against strong baselines in Table~\ref{tab:main_results}. Across all four benchmarks, our method consistently achieves superior performance, underscoring the strength of our routing framework. Several additional insights emerge. First, AgentRouter not only surpasses prior SOTA methods, but also outperforms the best individual agent on each benchmark (Best Agent baseline). This highlights that a collaboration of heterogeneous agents can effectively integrate complementary strengths, yielding higher accuracy than any single agent alone—demonstrating the necessity of adaptive routing. At the same time, the gap to the Oracle remains substantial, suggesting that there is still significant headroom and motivating further exploration of agent routing strategies.

Second, restricting the router’s output to the top-$K$ agents ($K{=}3,5,10$) unexpectedly produces the strongest performance, in some cases exceeding the unrestricted ensemble. A plausible explanation is that pruning the long tail of low-quality or noisy agents reduces variance and sharpens the aggregation of useful reasoning patterns. This suggests that agent routing benefits not only from diversity but also from judicious selection, where a smaller yet high-quality subset provides a better balance between complementarity and noise.

Finally, when comparing across benchmarks, we observe consistent gains on both multi-hop (2Wiki, HotpotQA) and single-hop/direct QA tasks (NewsQA, TriviaQA). The improvements are especially pronounced on multi-hop datasets, where reasoning requires combining evidence from multiple entities, and the advantage of heterogeneous collaboration becomes more salient. By contrast, in direct QA, the gains are smaller but still evident, indicating that even factual retrieval questions benefit from contextualized agent routing. Taken together, these results show that AgentRouter generalizes robustly across QA paradigms, with strong benefits in complex multi-hop reasoning scenarios.

\begin{figure}[t]
	\centering
	\includegraphics[width=1\linewidth]{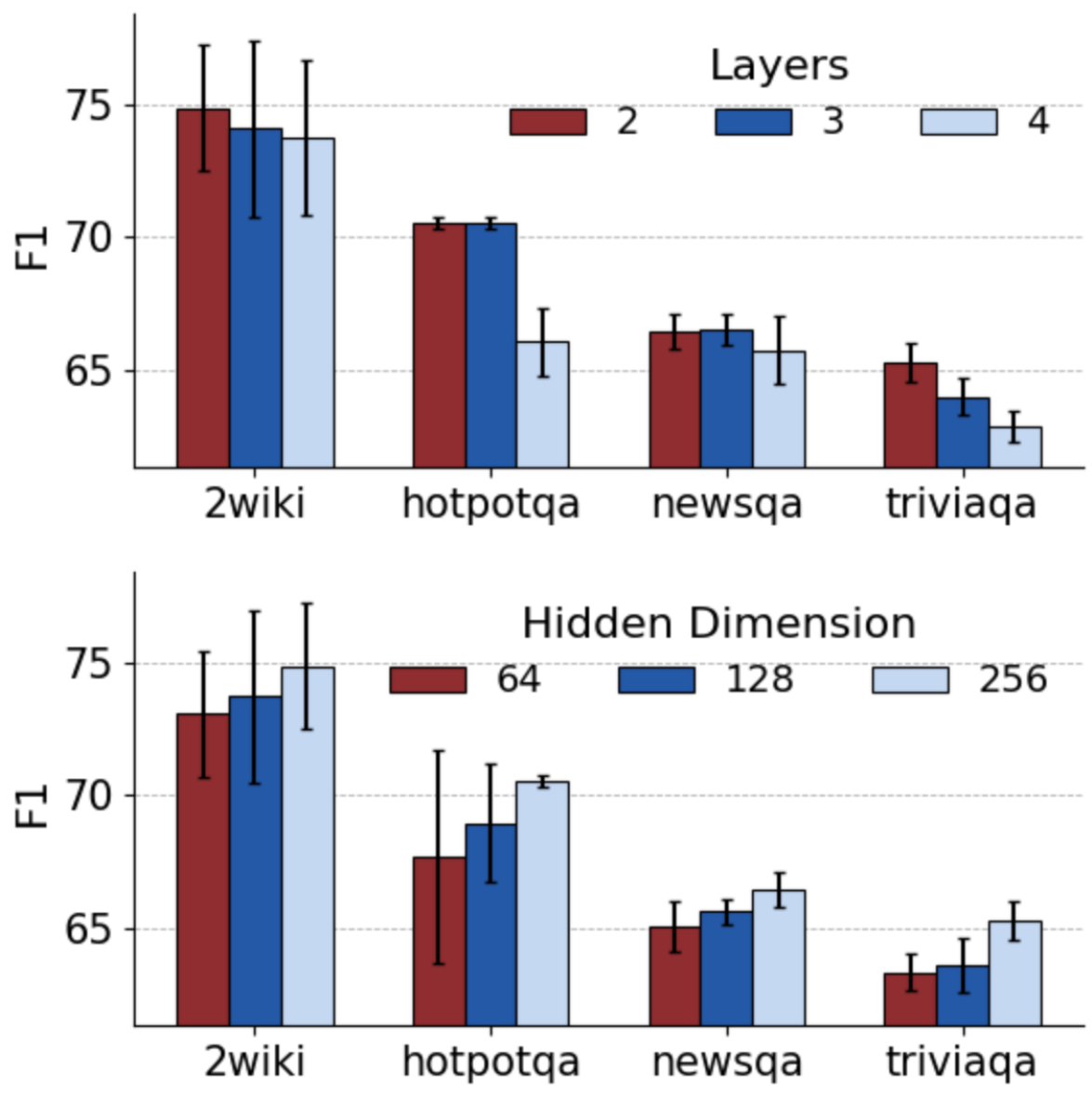}
        \vspace{-20pt}
	\caption{F1 performance on four QA benchmarks, varying Layers (top) and Hidden Dimensions (bottom). Error bars denote standard deviations.}
        \vspace{-15pt}
    \label{fig:hyper}
\end{figure}

\subsection{Ablation Analysis}

To better understand the effect of restricting the router’s output, we conduct an ablation study by varying the number of selected agents K. Figure~\ref{fig:topk} reports the relative change in F1 and EM, measured against the full ensemble (K${=}24$). Interestingly, we observe that performance does not monotonically increase with larger K. Instead, restricting the router to the top-K agents often yields different results on different benchmarks. Specifically, the clipping method usually works on multi-hop reasoning tasks, where improvements are most pronounced at moderate K (around $5$–$10$), while on single-hop QA tasks, usually more agents consistently yield better and more stable results. This trend might be due to some agents fail hard on multi-hop reasoning and removing the long tail of weaker agents reduces aggregation noise and variance, allowing more reliable reasoning signals to dominate. Overall, this ablation highlights that agent routing benefits from careful selection rather than unbounded inclusion on multi-hop complex reasoning. A smaller but higher-quality subset of agents provides a more stable and effective collaboration mechanism, underscoring the importance of top-K clipping in practice. 

We also study the effect of hyper-parameters. For example, Figure~\ref{fig:hyper} shows the analysis of the effect of hidden dimensions and layers. Across benchmarks, we observe that performance differences between configurations are modest but consistent. Increasing the number of layers does not always improve F1, with deeper settings sometimes yielding lower scores, particularly on multi-hop datasets. In contrast, larger hidden dimensions tend to offer more stable gains across tasks, suggesting that representational capacity plays a more robust role than network depth in this setting. These results motivate a closer examination of architectural trade-offs when designing models for diverse QA benchmarks. Similarly, we also conduct a series of other analyses, including the report of the original single agent behaviors on each task. We include these details in Appendix~\ref{appendix:additional}.

\subsection{Transferability Analysis}

As discussed in the previous section, agents built on the same backbone often exhibit wide and non-overlapping performance distributions, and their relative ranking varies substantially across tasks. We now examine this phenomenon from the perspective of \emph{transferability of the learned router weights}. Table~\ref{tab:transfer_2wiki}, \ref{tab:transfer_trivia} report results when the router is trained on one dataset and evaluated on another.

Several observations emerge. First, when the router is trained and tested on different tasks, performance generally deteriorates—particularly under strict Top-K clipping—demonstrating that the router assigns high weights to different subsets of agents depending on the task. This highlights the task-specific nature of agent specialization and reinforces the importance of adaptive routing. Second, the magnitude of degradation decreases as more agents are included, since adding a broader pool of candidates partially compensates for task mismatch. Third, transferability is not uniform across benchmarks: when the router is trained on TriviaQA (single-hop factual QA) and tested on 2Wiki (multi-hop reasoning), performance drops sharply, in some cases falling below simple heuristic ensembles. Conversely, training on 2Wiki and testing on TriviaQA also incurs a gap, but the effect is smaller. This asymmetry suggests that skills learned for multi-hop reasoning might be helpful to factual QA, whereas factual QA training fails to equip the router with strategies for complex reasoning.

Overall, these results underscore a key insight: \textsc{AgentRouter} is most effective when trained with task-relevant information, and it is necessary for an agent router to learn fine-grained, task-aware collaboration patterns. Unlike simple heuristic methods, our framework dynamically adapts to the demands of each benchmark, yielding consistently superior performance.  

\begin{table}[t]
\centering
\resizebox{\columnwidth}{!}{
\begin{tabular}{c| *{4}{c}}
\toprule
\multirow{2}{*}{\textbf{Top-K}} &
\multicolumn{2}{c}{\textbf{2Wiki} $\rightarrow$ \textbf{HotpotQA}} &
\multicolumn{2}{c}{\textbf{2Wiki} $\rightarrow$ \textbf{TriviaQA}} \\
\cmidrule(lr){2-3}\cmidrule(lr){4-5}
& F1 Drop\% & EM Drop\% & F1 Drop\% & EM Drop\% \\
\midrule
3  & 0.04  & -0.61 & 11.72 & 10.94 \\
6  & 0.00  &  0.00 & 11.97 & 14.88 \\
9  & -0.24 &  0.00 & 13.62 & 17.61 \\
12 & 0.43  &  1.15 & 14.73 & 17.45 \\
15 & 0.58  &  1.15 & 11.80 & 13.61 \\
18 & 1.66  &  2.32 & 11.70 & 14.06 \\
21 & 1.98  &  2.32 & 14.09 & 15.91 \\
24 & 1.98  &  2.32 & 11.87 & 13.64 \\
\bottomrule
\end{tabular}
}
\vspace{-5pt}
\caption{Performance drops when training on 2Wiki.}
\label{tab:transfer_2wiki}
\vspace{-5pt}
\end{table}

\begin{table}[t]
\centering
\resizebox{\columnwidth}{!}{
\begin{tabular}{c| *{4}{c}}
\toprule
\multirow{2}{*}{\textbf{Top-K}} &
\multicolumn{2}{c}{\textbf{TriviaQA} $\rightarrow$ \textbf{2Wiki}} &
\multicolumn{2}{c}{\textbf{TriviaQA} $\rightarrow$ \textbf{NewsQA}} \\
\cmidrule(lr){2-3}\cmidrule(lr){4-5}
& F1 Drop\% & EM Drop\% & F1 Drop\% & EM Drop\% \\
\midrule
3  & 35.04 & 41.94 & 17.53 & 42.67 \\
6  & 37.84 & 47.85 & 16.99 & 40.76 \\
9  & 32.07 & 40.75 & 15.08 & 37.50 \\
12 & 31.37 & 39.15 & 10.52 & 34.00 \\
15 & 30.91 & 38.97 &  7.39 & 26.49 \\
18 & 28.62 & 36.90 &  7.76 & 23.12 \\
21 & 29.07 & 38.16 &  6.66 & 21.25 \\
24 & 29.41 & 38.36 &  6.08 & 19.87 \\
\bottomrule
\end{tabular}
}
\vspace{-5pt}
\caption{Performance drops when training on TriviaQA.}
\label{tab:transfer_trivia}
\vspace{-15pt}
\end{table}

\section{Conclusion}
In this paper, we introduced \textsc{AgentRouter}, a knowledge-graph–guided agent routing framework. By modeling contexts and agents in a unified KG and learning task-aware collaboration strategies, our method consistently surpasses strong baselines across diverse QA benchmarks. Importantly, our results demonstrate that explicitly learning contextual information is critical: by guiding routing decisions with contextual information and supervised signals, \textsc{AgentRouter} is able to capture subtle task-dependent signals that simple heuristics miss, shedding light on new research paths. 


\newpage

\section*{Limitations}
While our study provides compelling evidence of the effectiveness of graph-supervised multi-agent routing, two limitations merit discussion. First, we did not explicitly address cost–performance trade-offs. Many routing approaches emphasize efficiency and cost reduction, whereas our design prioritizes performance and robustness. This choice is deliberate: our results show that \textsc{AgentRouter} achieves substantially higher accuracy than prior methods, highlighting the value of maximizing performance as a complementary perspective. Nevertheless, future work could incorporate cost-aware objectives to achieve more balanced trade-offs between efficiency and accuracy.  

Second, our framework currently operates on a fixed pool of manually designed agents. Although this captures diverse and representative strategies, it inherently limits exploration of the broader design space. Recent advances in automated agent design suggest that infinitely many agent variants could be generated and adapted dynamically. Integrating such automated generation into our routing framework represents a promising direction for future work, enabling continuous expansion of agent diversity. Importantly, our current results indicate that even within a limited agent pool, \textsc{AgentRouter} already achieves near-optimal performance relative to strong baselines, providing a solid foundation upon which more flexible and scalable extensions can be built.

\newpage
\newpage 
\bibliography{reference}

\begin{thebibliography}{69}
\providecommand{\natexlab}[1]{#1}

\bibitem[{Alibaba(2025)}]{qwen2025qwen2_5_7binstruct}
Qwen Alibaba. 2025.
\newblock Qwen2.5-7b-instruct.
\newblock \url{https://huggingface.co/Qwen/Qwen2.5-7B-Instruct}.

\bibitem[{Cao et~al.(2023)Cao, Xu, Yang, Wang, Zhang, Wang, Chen, and Yang}]{cao2023pre}
Yuxuan Cao, Jiarong Xu, Carl Yang, Jiaan Wang, Yunchao Zhang, Chunping Wang, Lei Chen, and Yang Yang. 2023.
\newblock When to pre-train graph neural networks? from data generation perspective!
\newblock In \emph{KDD}.

\bibitem[{Chen et~al.(2024{\natexlab{a}})Chen, Saha, and Bansal}]{chen2024reconcile}
Justin Chih-Yao Chen, Swarnadeep Saha, and Mohit Bansal. 2024{\natexlab{a}}.
\newblock Reconcile: Round-table conference improves reasoning via consensus among diverse {LLM}s.
\newblock In \emph{ACL}.

\bibitem[{Chen et~al.(2024{\natexlab{b}})Chen, Jiang, Lin, Kwok, and Zhang}]{chen2024routerdc}
Shuhao Chen, Weisen Jiang, Baijiong Lin, James Kwok, and Yu~Zhang. 2024{\natexlab{b}}.
\newblock Routerdc: Query-based router by dual contrastive learning for assembling large language models.
\newblock \emph{NeurIPS}.

\bibitem[{Chen et~al.(2024{\natexlab{c}})Chen, Su, Zuo, Yang, Yuan, Chan, Yu, Lu, Hung, Qian et~al.}]{chen2024agentverse}
Weize Chen, Yusheng Su, Jingwei Zuo, Cheng Yang, Chenfei Yuan, Chi-Min Chan, Heyang Yu, Yaxi Lu, Yi-Hsin Hung, Chen Qian, et~al. 2024{\natexlab{c}}.
\newblock Agentverse: Facilitating multi-agent collaboration and exploring emergent behaviors.
\newblock In \emph{ICLR}.

\bibitem[{Ding et~al.(2025)Ding, Mallick, Zhang, Wang et~al.}]{ding2025best}
D.~Ding, Ankur Mallick, Shaokun Zhang, Chi Wang, et~al. 2025.
\newblock Best-route: Adaptive llm routing with test-time optimal compute.
\newblock In \emph{ICML}.

\bibitem[{Ding et~al.(2024)Ding, Mallick, Wang, Sim, Mukherjee, Ruhle, Lakshmanan, and Awadallah}]{ding2024hybridllm}
Dujian Ding, Ankur Mallick, Chi Wang, Robert Sim, Subhabrata Mukherjee, Victor Ruhle, Laks~VS Lakshmanan, and Ahmed~Hassan Awadallah. 2024.
\newblock Hybridllm: Cost-efficient and quality-aware query routing.
\newblock In \emph{ICLR}.

\bibitem[{Du et~al.(2024)Du, Li, Torralba, Tenenbaum, and Mordatch}]{du2024mad}
Yilun Du, Shuang Li, Antonio Torralba, Joshua~B. Tenenbaum, and Igor Mordatch. 2024.
\newblock Improving factuality and reasoning in language models through multi-agent debate.
\newblock In \emph{ICML}.

\bibitem[{Fatemi et~al.(2023)Fatemi, Halcrow, and Perozzi}]{fatemi2023talk}
Bahare Fatemi, Jonathan Halcrow, and Bryan Perozzi. 2023.
\newblock Talk like a graph: Encoding graphs for large language models.
\newblock \emph{arXiv}.

\bibitem[{Feng et~al.(2025)Feng, Zhao, Liu, Yang, and Zhao}]{feng2024graphrouter}
Yihan Feng, Tianyu Zhao, Haotian Liu, Diyi Yang, and Tuo Zhao. 2025.
\newblock Graphrouter: Learning graph-based routing for large language model selection.
\newblock In \emph{ICLR}.

\bibitem[{Gao et~al.(2024)Gao, Qiao, Kan, Wen, He, and Li}]{gao2024two}
Yifu Gao, Linbo Qiao, Zhigang Kan, Zhihua Wen, Yongquan He, and Dongsheng Li. 2024.
\newblock Two-stage generative question answering on temporal knowledge graph using large language models.
\newblock \emph{arXiv}.

\bibitem[{Guo et~al.(2024{\natexlab{a}})Guo, Deng, Wen, Chen, Chang, and Wang}]{guo2024ds}
Siyuan Guo, Cheng Deng, Ying Wen, Hechang Chen, Yi~Chang, and Jun Wang. 2024{\natexlab{a}}.
\newblock Ds-agent: automated data science by empowering large language models with case-based reasoning.
\newblock In \emph{ICML}.

\bibitem[{Guo et~al.(2024{\natexlab{b}})Guo, Yang, Wang, Liu, Li, Tang, Li, and Wen}]{guo2024knowledgenavigator}
Tiezheng Guo, Qingwen Yang, Chen Wang, Yanyi Liu, Pan Li, Jiawei Tang, Dapeng Li, and Yingyou Wen. 2024{\natexlab{b}}.
\newblock Knowledgenavigator: Leveraging large language models for enhanced reasoning over knowledge graph.
\newblock \emph{Complex \& Intelligent Systems}.

\bibitem[{Hamilton et~al.(2017)Hamilton, Ying, and Leskovec}]{hamilton2017inductive}
Will Hamilton, Zhitao Ying, and Jure Leskovec. 2017.
\newblock Inductive representation learning on large graphs.
\newblock \emph{NeurIPS}.

\bibitem[{He et~al.(2024)He, Tian, Sun, Chawla, Laurent, LeCun, Bresson, and Hooi}]{he2024g}
Xiaoxin He, Yijun Tian, Yifei Sun, Nitesh~V Chawla, Thomas Laurent, Yann LeCun, Xavier Bresson, and Bryan Hooi. 2024.
\newblock G-retriever: Retrieval-augmented generation for textual graph understanding and question answering.
\newblock \emph{arXiv}.

\bibitem[{Ho et~al.(2020)Ho, Nguyen, Sugawara, and Aizawa}]{ho20202wikimultihopqa}
Xanh Ho, Anh-Khoa~Duong Nguyen, Saku Sugawara, and Akiko Aizawa. 2020.
\newblock Constructing a multi-hop qa dataset for comprehensive evaluation of reasoning steps.
\newblock In \emph{COLING}.

\bibitem[{Jiang et~al.(2023{\natexlab{a}})Jiang, Ren, and Lin}]{jiang2023llm}
Dongfu Jiang, Xiang Ren, and Bill~Yuchen Lin. 2023{\natexlab{a}}.
\newblock {LLM}-blender: Ensembling large language models with pairwise ranking and generative fusion.
\newblock In \emph{ACL}.

\bibitem[{Jiang et~al.(2023{\natexlab{b}})Jiang, Zhou, Dong, Ye, Zhao, and Wen}]{jiang2023structgpt}
Jinhao Jiang, Kun Zhou, Zican Dong, Keming Ye, Wayne~Xin Zhao, and Ji-Rong Wen. 2023{\natexlab{b}}.
\newblock Structgpt: A general framework for large language model to reason over structured data.
\newblock In \emph{EMNLP}.

\bibitem[{Jiang et~al.(2024)Jiang, Zhou, Zhao, Song, Zhu, Zhu, and Wen}]{jiang2024kg}
Jinhao Jiang, Kun Zhou, Wayne~Xin Zhao, Yang Song, Chen Zhu, Hengshu Zhu, and Ji-Rong Wen. 2024.
\newblock Kg-agent: An efficient autonomous agent framework for complex reasoning over knowledge graph.
\newblock \emph{arXiv}.

\bibitem[{Jimenez et~al.(2024)Jimenez, Yang, Wettig, Yao, Pei, Press, and Narasimhan}]{jimenez2024swe}
Carlos~E Jimenez, John Yang, Alexander Wettig, Shunyu Yao, Kexin Pei, Ofir Press, and Karthik~R Narasimhan. 2024.
\newblock Swe-bench: Can language models resolve real-world github issues?
\newblock In \emph{ICLR}.

\bibitem[{Jin et~al.(2024)Jin, Xie, Zhang, Roy, Zhang, Li, Li, Tang, Wang, Meng et~al.}]{jin2024graph}
Bowen Jin, Chulin Xie, Jiawei Zhang, Kashob~Kumar Roy, Yu~Zhang, Zheng Li, Ruirui Li, Xianfeng Tang, Suhang Wang, Yu~Meng, et~al. 2024.
\newblock Graph chain-of-thought: Augmenting large language models by reasoning on graphs.
\newblock In \emph{ACL}.

\bibitem[{Joshi et~al.(2017)Joshi, Choi, Weld, and Zettlemoyer}]{joshi2017triviaqa}
Mandar Joshi, Eunsol Choi, Daniel Weld, and Luke Zettlemoyer. 2017.
\newblock Triviaqa: A large scale distantly supervised challenge dataset for reading comprehension.
\newblock In \emph{ACL}.

\bibitem[{Kim et~al.(2023)Kim, Kwon, Jo, and Choi}]{kim2023kg}
Jiho Kim, Yeonsu Kwon, Yohan Jo, and Edward Choi. 2023.
\newblock Kg-gpt: A general framework for reasoning on knowledge graphs using large language models.
\newblock In \emph{EMNLP}.

\bibitem[{Kipf and Welling(2016)}]{kipf2016semi}
Thomas~N Kipf and Max Welling. 2016.
\newblock Semi-supervised classification with graph convolutional networks.
\newblock \emph{arXiv}.

\bibitem[{Kojima et~al.(2022)Kojima, Gu, Reid, Matsuo, and Iwasawa}]{kojima2022large}
Takeshi Kojima, Shixiang~Shane Gu, Machel Reid, Yutaka Matsuo, and Yusuke Iwasawa. 2022.
\newblock Large language models are zero-shot reasoners.
\newblock \emph{NeurIPS}.

\bibitem[{Lazaridou et~al.(2022)Lazaridou, Gribovskaya, Stokowiec, and Grigorev}]{lazaridou2022internet}
Angeliki Lazaridou, Elena Gribovskaya, Wojciech Stokowiec, and Nikolai Grigorev. 2022.
\newblock Internet-augmented language models through few-shot prompting for open-domain question answering.
\newblock \emph{arXiv}.

\bibitem[{Lewis et~al.(2020)Lewis, Perez, Piktus, Petroni, Karpukhin, Goyal, K{\"u}ttler, Lewis, Yih, Rockt{\"a}schel et~al.}]{lewis2020retrieval}
Patrick Lewis, Ethan Perez, Aleksandra Piktus, Fabio Petroni, Vladimir Karpukhin, Naman Goyal, Heinrich K{\"u}ttler, Mike Lewis, Wen-tau Yih, Tim Rockt{\"a}schel, et~al. 2020.
\newblock Retrieval-augmented generation for knowledge-intensive nlp tasks.
\newblock \emph{NeuralIPS}.

\bibitem[{Li et~al.(2024)Li, Zhao, Wang, Wang, Zhou, Srivastava, Gokmen, Lee, Li, Zhang et~al.}]{li2024embodied}
Manling Li, Shiyu Zhao, Qineng Wang, Kangrui Wang, Yu~Zhou, Sanjana Srivastava, Cem Gokmen, Tony Lee, Li~Erran Li, Ruohan Zhang, et~al. 2024.
\newblock Embodied agent interface: benchmarking llms for embodied decision making.
\newblock In \emph{NeuralIPS}.

\bibitem[{Liu et~al.(2024)Liu, He, Tian, and Chawla}]{liu2024can}
Zheyuan Liu, Xiaoxin He, Yijun Tian, and Nitesh~V Chawla. 2024.
\newblock Can we soft prompt llms for graph learning tasks?
\newblock In \emph{WWW}, pages 481--484.

\bibitem[{Ma et~al.(2025{\natexlab{a}})Ma, Qian, Wang, Zhang, Zhang, and Ye}]{ma2025llm}
Tianyi Ma, Yiyue Qian, Zehong Wang, Zheyuan Zhang, Chuxu Zhang, and Yanfang Ye. 2025{\natexlab{a}}.
\newblock Llm-empowered class imbalanced graph prompt learning for online drug trafficking detection.
\newblock \emph{arXiv}.

\bibitem[{Ma et~al.(2023)Ma, Qian, Zhang, and Ye}]{ma2023hypergraph}
Tianyi Ma, Yiyue Qian, Chuxu Zhang, and Yanfang Ye. 2023.
\newblock Hypergraph contrastive learning for drug trafficking community detection.
\newblock In \emph{ICDM}.

\bibitem[{Ma et~al.(2025{\natexlab{b}})Ma, Qian, Zhang, Zhang, and Ye}]{ma2025adaptive}
Tianyi Ma, Yiyue Qian, Shinan Zhang, Chuxu Zhang, and Yanfang Ye. 2025{\natexlab{b}}.
\newblock Adaptive expansion for hypergraph learning.
\newblock \emph{arXiv preprint arXiv:2502.15564}.

\bibitem[{Ma et~al.(2025{\natexlab{c}})Ma, Qian, Zhang, Wang, Qian, Bai, Ding, Luo, Zhang, Murugesan et~al.}]{ma2025autodata}
Tianyi Ma, Yiyue Qian, Zheyuan Zhang, Zehong Wang, Xiaoye Qian, Feifan Bai, Yifan Ding, Xuwei Luo, Shinan Zhang, Keerthiram Murugesan, et~al. 2025{\natexlab{c}}.
\newblock Autodata: A multi-agent system for open web data collection.
\newblock \emph{arXiv preprint arXiv:2505.15859}.

\bibitem[{MetaAI(2024)}]{meta2024llama3}
MetaAI. 2024.
\newblock Llama-3-8b-instruct.
\newblock \url{https://huggingface.co/meta-llama/Llama-3-8B-Instruct}.

\bibitem[{MistralAI(2023)}]{mistral2023mixtral_8x7binstruct}
MistralAI. 2023.
\newblock Mixtral-8x7b-instruct-v0.1.
\newblock \url{https://huggingface.co/mistralai/Mixtral-8x7B-Instruct-v0.1}.

\bibitem[{Ni et~al.(2025)Ni, Liu, Wang, Lei, Zhao, Cheng, Zeng, Dong, Xia, Kenthapadi et~al.}]{ni2025towards}
Bo~Ni, Zheyuan Liu, Leyao Wang, Yongjia Lei, Yuying Zhao, Xueqi Cheng, Qingkai Zeng, Luna Dong, Yinglong Xia, Krishnaram Kenthapadi, et~al. 2025.
\newblock Towards trustworthy retrieval augmented generation for large language models: A survey.
\newblock \emph{arXiv}.

\bibitem[{Ong et~al.(2025)Ong, Almahairi, Wu, Chiang, Wu, Gonzalez, Kadous, and Stoica}]{ong2024routellm}
Isaac Ong, Amjad Almahairi, Vincent Wu, Wei‐Lin Chiang, Tianhao Wu, Joseph~E. Gonzalez, M.~Waleed Kadous, and Ion Stoica. 2025.
\newblock Routellm: Learning to route llms with preference data.
\newblock In \emph{ICLR}.

\bibitem[{OpenAI(2025)}]{gptoss2025gpt_oss_20b}
OpenAI. 2025.
\newblock Gpt-oss-20b.
\newblock \url{https://llm-stats.com/models/gpt-oss-20b}.

\bibitem[{Ouyang et~al.(2022)Ouyang, Wu, Jiang, Almeida, Wainwright, Mishkin, Zhang, Agarwal, Slama, Ray et~al.}]{ouyang2022training}
Long Ouyang, Jeffrey Wu, Xu~Jiang, Diogo Almeida, Carroll Wainwright, Pamela Mishkin, Chong Zhang, Sandhini Agarwal, Katarina Slama, Alex Ray, et~al. 2022.
\newblock Training language models to follow instructions with human feedback.
\newblock \emph{NeuralIPS}.

\bibitem[{Peng et~al.(2024)Peng, Zhu, Liu, Bo, Shi, Hong, Zhang, and Tang}]{peng2024graph}
Boci Peng, Yun Zhu, Yongchao Liu, Xiaohe Bo, Haizhou Shi, Chuntao Hong, Yan Zhang, and Siliang Tang. 2024.
\newblock Graph retrieval-augmented generation: A survey.
\newblock \emph{arXiv}.

\bibitem[{Qian et~al.(2024)Qian, Ma, Zhang, and Ye}]{qian2024dual}
Yiyue Qian, Tianyi Ma, Chuxu Zhang, and Yanfang Ye. 2024.
\newblock Dual-level hypergraph contrastive learning with adaptive temperature enhancement.
\newblock In \emph{WWW}.

\bibitem[{Rajpurkar et~al.(2016)Rajpurkar, Zhang, Lopyrev, and Liang}]{rajpurkar2016squad}
Pranav Rajpurkar, Jian Zhang, Konstantin Lopyrev, and Percy Liang. 2016.
\newblock Squad: 100,000+ questions for machine comprehension of text.
\newblock In \emph{EMNLP}.

\bibitem[{Shinn et~al.(2023)Shinn, Cassano, Gopinath, Narasimhan, and Yao}]{shinn2023reflexion}
Noah Shinn, Federico Cassano, Ashwin Gopinath, Karthik Narasimhan, and Shunyu Yao. 2023.
\newblock Reflexion: Language agents with verbal reinforcement learning.
\newblock \emph{NeurIPS}.

\bibitem[{Singh et~al.(2025)Singh, Ehtesham, Kumar, and Khoei}]{singh2025agentic}
Aditi Singh, Abul Ehtesham, Saket Kumar, and Tala~Talaei Khoei. 2025.
\newblock Agentic retrieval-augmented generation: A survey on agentic rag.
\newblock \emph{arXiv}.

\bibitem[{Stripelis et~al.(2024)}]{stripelis2024multi}
D.~Stripelis et~al. 2024.
\newblock A multi-model router for efficient llm inference.
\newblock In \emph{EMNLP}.

\bibitem[{Sun et~al.(2019)Sun, Bedrax-Weiss, and Cohen}]{sun2019pullnet}
Haitian Sun, Tania Bedrax-Weiss, and William Cohen. 2019.
\newblock Pullnet: Open domain question answering with iterative retrieval on knowledge bases and text.
\newblock In \emph{EMNLP-IJCNLP}.

\bibitem[{Taunk et~al.(2023)Taunk, Khanna, Kandru, Varma, Sharma, and Tapaswi}]{taunk2023grapeqa}
Dhaval Taunk, Lakshya Khanna, Siri Venkata Pavan~Kumar Kandru, Vasudeva Varma, Charu Sharma, and Makarand Tapaswi. 2023.
\newblock Grapeqa: Graph augmentation and pruning to enhance question-answering.
\newblock In \emph{WWW}.

\bibitem[{Trischler et~al.(2017)Trischler, Wang, Yuan, Harris, Sordoni, Bachman, and Suleman}]{trischler2017newsqa}
Adam Trischler, Tong Wang, Xingdi Yuan, Justin Harris, Alessandro Sordoni, Philip Bachman, and Kaheer Suleman. 2017.
\newblock Newsqa: A machine comprehension dataset.
\newblock In \emph{RepL4NLP}.

\bibitem[{Veli{\v{c}}kovi{\'c} et~al.(2017)Veli{\v{c}}kovi{\'c}, Cucurull, Casanova, Romero, Lio, and Bengio}]{velivckovic2017graph}
Petar Veli{\v{c}}kovi{\'c}, Guillem Cucurull, Arantxa Casanova, Adriana Romero, Pietro Lio, and Yoshua Bengio. 2017.
\newblock Graph attention networks.
\newblock \emph{arXiv}.

\bibitem[{Wang et~al.(2024{\natexlab{a}})Wang, Feng, He, Tan, Han, and Tsvetkov}]{wang2024can}
Heng Wang, Shangbin Feng, Tianxing He, Zhaoxuan Tan, Xiaochuang Han, and Yulia Tsvetkov. 2024{\natexlab{a}}.
\newblock Can language models solve graph problems in natural language?
\newblock \emph{NeuralIPS}.

\bibitem[{Wang et~al.(2025{\natexlab{a}})Wang, Fu, Zhang, Cheng et~al.}]{wang2025mixllm}
X.~Wang, Y.~Fu, Y.~Zhang, W.~Cheng, et~al. 2025{\natexlab{a}}.
\newblock Mixllm: Dynamic routing in mixed large language models.
\newblock In \emph{NAACL}.

\bibitem[{Wang et~al.(2023{\natexlab{a}})Wang, Yang, Qiu, Liang, He, Gu, Xiao, and Wang}]{wang2023knowledgpt}
Xintao Wang, Qianwen Yang, Yongting Qiu, Jiaqing Liang, Qianyu He, Zhouhong Gu, Yanghua Xiao, and Wei Wang. 2023{\natexlab{a}}.
\newblock Knowledgpt: Enhancing large language models with retrieval and storage access on knowledge bases.
\newblock \emph{arXiv}.

\bibitem[{Wang et~al.(2023{\natexlab{b}})Wang, Wei, Schuurmans, Le, Chi, and Zhou}]{wang2023selfconsistency}
Xuezhi Wang, Jason Wei, Dale Schuurmans, Quoc Le, Ed~Chi, and Denny Zhou. 2023{\natexlab{b}}.
\newblock Self-consistency improves chain of thought reasoning in language models.
\newblock In \emph{ICLR}.

\bibitem[{Wang et~al.(2022)Wang, Wei, Schuurmans, Le, Chi, Narang, Chowdhery, and Zhou}]{wang2022self}
Xuezhi Wang, Jason Wei, Dale Schuurmans, Quoc~V Le, Ed~H Chi, Sharan Narang, Aakanksha Chowdhery, and Denny Zhou. 2022.
\newblock Self-consistency improves chain of thought reasoning in language models.
\newblock In \emph{ICLR}.

\bibitem[{Wang et~al.(2025{\natexlab{b}})Wang, Liu, Zhang, Ma, Zhang, and Ye}]{wang2025can}
Zehong Wang, Sidney Liu, Zheyuan Zhang, Tianyi Ma, Chuxu Zhang, and Yanfang Ye. 2025{\natexlab{b}}.
\newblock Can llms convert graphs to text-attributed graphs?
\newblock In \emph{NAACL}.

\bibitem[{Wang et~al.(2024{\natexlab{b}})Wang, Zhang, Chawla, Zhang, and Ye}]{wang2024gft}
Zehong Wang, Zheyuan Zhang, Nitesh Chawla, Chuxu Zhang, and Yanfang Ye. 2024{\natexlab{b}}.
\newblock Gft: Graph foundation model with transferable tree vocabulary.
\newblock \emph{NeruIPS}.

\bibitem[{Wang et~al.(2025{\natexlab{c}})Wang, Zhang, Ma, Chawla, Zhang, and Ye}]{wang2024learning}
Zehong Wang, Zheyuan Zhang, Tianyi Ma, Nitesh~V Chawla, Chuxu Zhang, and Yanfang Ye. 2025{\natexlab{c}}.
\newblock Learning cross-task generalities across graphs via task-trees.
\newblock \emph{ICML}.

\bibitem[{Wang et~al.(2025{\natexlab{d}})Wang, Zhang, Ma, Chawla, Zhang, and Ye}]{wang2025neural}
Zehong Wang, Zheyuan Zhang, Tianyi Ma, Nitesh~V Chawla, Chuxu Zhang, and Yanfang Ye. 2025{\natexlab{d}}.
\newblock Neural graph pattern machine.
\newblock \emph{ICML}.

\bibitem[{Wang et~al.(2024{\natexlab{c}})Wang, Zhang, Zhang, and Ye}]{wang2024subgraph}
Zehong Wang, Zheyuan Zhang, Chuxu Zhang, and Yanfang Ye. 2024{\natexlab{c}}.
\newblock Subgraph pooling: tackling negative transfer on graphs.
\newblock In \emph{IJCAI}.

\bibitem[{Wei et~al.(2022)Wei, Wang, Schuurmans, Bosma, Xia, Chi, Le, Zhou et~al.}]{wei2022chain}
Jason Wei, Xuezhi Wang, Dale Schuurmans, Maarten Bosma, Fei Xia, Ed~Chi, Quoc~V Le, Denny Zhou, et~al. 2022.
\newblock Chain-of-thought prompting elicits reasoning in large language models.
\newblock \emph{NeuralIPS}.

\bibitem[{Wen et~al.(2023)Wen, Wang, and Sun}]{wen2023mindmap}
Yilin Wen, Zifeng Wang, and Jimeng Sun. 2023.
\newblock Mindmap: Knowledge graph prompting sparks graph of thoughts in large language models.
\newblock \emph{arXiv}.

\bibitem[{Yang et~al.(2018)Yang, Qi, Zhang, Bengio, Cohen, Salakhutdinov, and Manning}]{yang2018hotpotqa}
Zhilin Yang, Peng Qi, Saizheng Zhang, Yoshua Bengio, William~W. Cohen, Ruslan Salakhutdinov, and Christopher~D. Manning. 2018.
\newblock Hotpotqa: A dataset for diverse, explainable multi-hop question answering.
\newblock In \emph{EMNLP}.

\bibitem[{Yao et~al.(2023)Yao, Zhao, Yu, Du, Shafran, Narasimhan, and Cao}]{yao2023react}
Shunyu Yao, Jeffrey Zhao, Dian Yu, Nan Du, Izhak Shafran, Karthik Narasimhan, and Yuan Cao. 2023.
\newblock React: Synergizing reasoning and acting in language models.
\newblock In \emph{ICLR}.

\bibitem[{Yasunaga et~al.(2021)Yasunaga, Ren, Bosselut, Liang, and Leskovec}]{yasunaga2021qa}
Michihiro Yasunaga, Hongyu Ren, Antoine Bosselut, Percy Liang, and Jure Leskovec. 2021.
\newblock Qa-gnn: Reasoning with language models and knowledge graphs for question answering.
\newblock In \emph{NAACL}.

\bibitem[{Zhang et~al.(2022)Zhang, Zhang, Yu, Tang, Tang, Li, and Chen}]{zhang2022subgraph}
Jing Zhang, Xiaokang Zhang, Jifan Yu, Jian Tang, Jie Tang, Cuiping Li, and Hong Chen. 2022.
\newblock Subgraph retrieval enhanced model for multi-hop knowledge base question answering.
\newblock In \emph{ACL}.

\bibitem[{Zhang et~al.(2025{\natexlab{a}})Zhang, Li, Le, Wang, Ma, Galassi, Murugesan, Moniz, Geyer, Chawla et~al.}]{zhang2024ngqa}
Zheyuan Zhang, Yiyang Li, Nhi Ha~Lan Le, Zehong Wang, Tianyi Ma, Vincent Galassi, Keerthiram Murugesan, Nuno Moniz, Werner Geyer, Nitesh~V Chawla, et~al. 2025{\natexlab{a}}.
\newblock Ngqa: A nutritional graph question answering benchmark for personalized health-aware nutritional reasoning.
\newblock \emph{ACL}.

\bibitem[{Zhang et~al.(2025{\natexlab{b}})Zhang, Ma, Wang, Li, Hou, Sun, Shi, Ma, Song, Abbasi et~al.}]{zhang2025llms4all}
Zheyuan Zhang, Tianyi Ma, Zehong Wang, Yiyang Li, Shifu Hou, Weixiang Sun, Kaiwen Shi, Yijun Ma, Wei Song, Ahmed Abbasi, et~al. 2025{\natexlab{b}}.
\newblock Llms4all: A review on large language models for research and applications in academic disciplines.
\newblock \emph{arXiv preprint arXiv:2509.19580}.

\bibitem[{Zhang et~al.(2024{\natexlab{a}})Zhang, Wang, Hou, Hall, Bachman, White, Galassi, Chawla, Zhang, and Ye}]{zhang2024diet}
Zheyuan Zhang, Zehong Wang, Shifu Hou, Evan Hall, Landon Bachman, Jasmine White, Vincent Galassi, Nitesh~V Chawla, Chuxu Zhang, and Yanfang Ye. 2024{\natexlab{a}}.
\newblock Diet-odin: A novel framework for opioid misuse detection with interpretable dietary patterns.
\newblock In \emph{Proceedings of the 30th ACM SIGKDD Conference on Knowledge Discovery and Data Mining}, pages 6312--6323.

\bibitem[{Zhang et~al.(2024{\natexlab{b}})Zhang, Wang, Ma, Taneja, Nelson, Le, Murugesan, Ju, Chawla, Zhang et~al.}]{zhang2024mopi}
Zheyuan Zhang, Zehong Wang, Tianyi Ma, Varun~Sameer Taneja, Sofia Nelson, Nhi Ha~Lan Le, Keerthiram Murugesan, Mingxuan Ju, Nitesh~V Chawla, Chuxu Zhang, et~al. 2024{\natexlab{b}}.
\newblock Mopi-hfrs: A multi-objective personalized health-aware food recommendation system with llm-enhanced interpretation.
\newblock \emph{arXiv}.

\end{thebibliography}

\newpage
\appendix

\section{Related Works}
\subsection{Task-Adaptive Agent Selection}
Large language models (LLMs) and agentic frameworks have advanced rapidly in recent years \cite{zhang2025llms4all}. With the rise of LLM-based agents, studies increasingly show that no single LLM or agent uniformly excels across downstream tasks; rather, different models or agents exhibit \emph{complementary strengths}. For example, AgentVerse demonstrates that multi-agent groups can outperform individual agents and display emergent collaborative behaviors \citep{chen2024agentverse}, while ReConcile shows that organizing diverse LLMs into rounds of discussion with consensus voting yields superior reasoning compared to single-model baselines \citep{chen2024reconcile}. Given these findings, a natural next step is input-conditioned \emph{selection and coordination}—that is, learning to route or assemble specialized models and agents per query to harvest complementary gains. Early efforts pursued this idea through simple binary or ensemble-based collaboration: majority-vote aggregation (e.g., self-consistency \citep{wang2022self}) and static ensembling confirmed the benefit of diversity but fixed the model set in advance, offering little adaptivity to input characteristics and limited guidance on which agent should be prioritized for a given instance \citep{jiang2023llm}.  
Subsequent research introduced learned routers that predict which LLM is best suited for a query, thereby providing more adaptive selection than static ensembles. RouteLLM, for instance, learns routing from human preference data and generalizes across strong–weak model pairs \citep{ong2024routellm}. RouterDC employs dual contrastive learning to assemble multiple LLMs, achieving gains over top individual models both in-distribution and out-of-distribution \citep{chen2024routerdc}. MixLLM frames routing as a contextual bandit problem, enabling dynamic adaptation under evolving query distributions and mixed LLM pools \citep{wang2025mixllm}. In parallel, other advances emphasized coordination among specialized agents rather than selecting a single model. Approaches such as TO-Router and BEST-Route adaptively determine how many outputs to sample or which expert models to involve depending on query difficulty, rather than relying on fixed pipelines \citep{stripelis2024multi, ding2025best}. While these approaches highlight the value of adaptivity, many still rely on heuristic collaboration rules or shallow controllers, and they seldom capture richer inter-dependencies among queries, tasks, and agents.  

The most recent wave of research formulates routing as a structured learning problem. For example, \textsc{GraphRouter} casts routing as a link prediction task on a heterogeneous graph and leverages graph neural networks to jointly model query–model, query–query, and model–model relations \citep{feng2024graphrouter}. This represents a clear step beyond shallow controllers or pairwise scoring, yet it remains limited in its ability to incorporate fine-grained task semantics or supervised graph signals that can more directly guide adaptive collaboration across diverse agent designs. 

\subsection{Knowledge Graph Question Answering}
Research on Knowledge Graph Question Answering (KGQA) has progressed from classic semantic parsing and retrieval paradigms to increasingly model-driven solutions. Early systems translated natural-language questions into executable logical forms (e.g., SPARQL) over a knowledge graph \cite{sun2019pullnet, zhang2022subgraph}, often pairing pre-trained encoders such as BERT with graph-aware architectures (GNNs/LSTMs) to locate entities, relations, and supporting subgraphs \cite{yasunaga2021qa, taunk2023grapeqa}.  
More recent approaches incorporate large language models (LLMs) to improve both access and reasoning: some convert questions into structured queries like SQL/SPARQL to sharpen retrieval \cite{jiang2023structgpt, wang2023knowledgpt}, while others emphasize multi-hop inference over retrieved triples or subgraphs to handle compositional reasoning \cite{kim2023kg, gao2024two}. Despite these advances, widely used benchmarks remain largely general-purpose and do not fully capture domain-specific demands—e.g., the nuanced constraints present in nutritional-health reasoning scenarios.

\subsection{Graph-Retrieval Augmented Generation}
Graph-Retrieval Augmented Generation (Graph-RAG) generalizes the RAG paradigm \cite{lewis2020retrieval, ni2025towards} by retrieving \emph{structured} evidence rather than only unstructured text. Instead of passages alone, Graph-RAG surfaces graph fragments (triples/subgraphs) and uses graph encoders to condition generation, thereby improving precision and reducing redundancy \cite{guo2024knowledgenavigator, wen2023mindmap, lazaridou2022internet, liu2024can}.  
Current evaluations predominantly probe elementary graph reasoning skills—such as path finding, degree/counting, or edge existence \cite{fatemi2023talk, wang2024can, wang2025can}. While informative for fundamentals, these settings under-represent domain-specific requirements. He et al.\ introduce more advanced graph-understanding benchmarks in general contexts \cite{he2024g}, yet tailored evaluations for domains like nutrition remain scarce \cite{zhang2024mopi, zhang2024ngqa}. Building on Graph-RAG principles, many applications nowadays thrive and shed lights on new research paths \cite{zhang2024diet}

\subsection{Graph Neural Networks.}
Graph Neural Networks (GNNs) are designed for relational data and have delivered strong results across social, recommendation, biological, and molecular applications by exploiting graph inductive biases \cite{kipf2016semi, velivckovic2017graph, hamilton2017inductive}. Their ability to share parameters across varying graph sizes/topologies supports deployment in dynamic, real-world settings. A growing body of work investigates transfer and pretraining for cross-task/domain generalization—mirroring trends in language and vision—via subgraph pooling, pretraining schemes, and task-agnostic embeddings \cite{ma2023hypergraph, wang2024subgraph, ma2025llm, cao2023pre, ma2025adaptive}. Looking forward, the community is moving toward \emph{graph foundation models}, i.e., large-scale pretrained GNN backbones intended to capture broadly reusable structural/semantic patterns \cite{wang2024gft, qian2024dual, wang2025neural, wang2024learning}. Despite progress, open challenges persist, including over-smoothing, expressive-power limits, and scalability, motivating research on more adaptive architectures and training recipes.

\section{Implementation Details}
\subsection{Benchmarks}
\label{appendix:Benchmarks}

\textbf{HotpotQA}~\citep{yang2018hotpotqa} is a large-scale question answering benchmark explicitly designed to evaluate multi-hop reasoning. Unlike traditional QA datasets where answers can be located within a single passage, HotpotQA requires systems to combine information across multiple documents to arrive at the correct answer. It contains over 100,000 question–answer pairs derived from Wikipedia, annotated with supporting sentences that enable explainable reasoning. This benchmark challenges models not only to retrieve the right evidence but also to demonstrate the ability to integrate scattered information coherently. Given its emphasis on multi-step inference, HotpotQA serves as a critical testbed for evaluating reasoning beyond simple fact extraction. 

\noindent\textbf{2WikiMultihopQA}~\citep{ho20202wikimultihopqa} extends the idea of multi-hop reasoning by constructing questions that require reasoning across pairs of Wikipedia articles. Compared to HotpotQA, it provides more challenging scenarios where supporting evidence is deliberately spread across two distinct documents, forcing models to bridge semantic gaps between disparate sources. The dataset includes a diverse range of question types, from entity relations to compositional reasoning, thereby testing both retrieval and reasoning capabilities. By requiring models to locate and combine information from multiple, sometimes loosely connected passages, 2WikiMultihopQA offers a stringent benchmark for assessing deeper reasoning and robust evidence integration. 

\noindent\textbf{NewsQA}~\citep{trischler2017newsqa} is a large-scale question answering dataset constructed from CNN news articles. It consists of over 100,000 human-generated questions paired with answers derived from corresponding news passages. Unlike earlier QA datasets that focus on simple fact extraction, NewsQA emphasizes reasoning, inference, and synthesis across multiple sentences within an article. Its design introduces ambiguity, unanswerable questions, and multi-sentence reasoning, making it a challenging benchmark for evaluating reading comprehension and open-domain question answering systems. 

\noindent\textbf{TriviaQA}~\citep{joshi2017triviaqa} is a QA benchmark built from trivia-style questions collected from online sources. It contains over 95,000 question–answer pairs, with evidence passages drawn from Wikipedia and the broader web. TriviaQA is particularly challenging because its questions are authored independently of the supporting documents, resulting in diverse phrasing, lexical variation, and indirect evidence. This property forces models to rely on semantic understanding rather than surface-level matching. As such, TriviaQA tests both retrieval robustness and reasoning generalization under noisy, real-world conditions. 

\begin{table*}[t]
\centering
\resizebox{\textwidth}{!}{
\begin{tabular}{c c cccccc}
\toprule
\textbf{Benchmark} & \textbf{Agent} & \textbf{Raw} & \textbf{CoT} & \textbf{SC} & \textbf{React Reflect} & \textbf{MAD} & \textbf{Summary} \\
\midrule
\multirow{4}{*}{NewsQA} 
 & Qwen2.5-7B   & 60.42$\pm$1.74 & 59.83$\pm$2.63 & 58.91$\pm$1.62 & 59.95$\pm$1.75 & 59.94$\pm$2.81 & 60.27$\pm$1.50 \\
 & gpt-oss-20B  & 59.89$\pm$2.62 & 61.07$\pm$1.51 & 59.36$\pm$2.50 & 59.55$\pm$1.64 & 61.78$\pm$1.89 & 61.69$\pm$2.33 \\
 & Mixtral-8x7B & 54.12$\pm$1.35 & 53.75$\pm$2.37 & 56.10$\pm$1.67 & 54.68$\pm$1.46 & 54.96$\pm$2.75 & 55.51$\pm$2.84 \\
 & Llama-3-8B   & 55.24$\pm$1.99 & 56.42$\pm$1.25 & 58.26$\pm$2.45 & 56.82$\pm$2.24 & 56.76$\pm$1.86 & 55.73$\pm$1.57 \\
\midrule
\multirow{4}{*}{HotpotQA}
 & Qwen2.5-7B   & 59.69$\pm$1.21 & 57.73$\pm$1.32 & 58.32$\pm$1.37 & 57.20$\pm$1.60 & 55.89$\pm$2.48 & 56.93$\pm$2.72 \\
 & gpt-oss-20B  & 68.32$\pm$1.65 & 68.68$\pm$1.70 & 67.35$\pm$1.81 & 68.62$\pm$1.04 & 67.43$\pm$1.93 & 66.40$\pm$1.09 \\
 & Mixtral-8x7B & 58.63$\pm$0.47 & 57.23$\pm$0.58 & 53.49$\pm$0.70 & 58.45$\pm$0.86 & 58.57$\pm$0.75 & 58.75$\pm$0.98 \\
 & Llama-3-8B   & 56.50$\pm$1.02 & 54.95$\pm$1.14 & 56.06$\pm$1.26 & 56.31$\pm$1.42 & 54.74$\pm$1.37 & 52.32$\pm$1.53 \\
\midrule
\multirow{4}{*}{2Wiki}
 & Qwen2.5-7B   & 41.94$\pm$2.20 & 41.91$\pm$1.29 & 40.92$\pm$2.37 & 44.65$\pm$1.45 & 42.17$\pm$1.54 & 44.56$\pm$1.62 \\
 & gpt-oss-20B  & 71.27$\pm$1.71 & 70.63$\pm$1.79 & 71.28$\pm$1.87 & 73.80$\pm$1.96 & 70.50$\pm$2.04 & 74.89$\pm$1.12 \\
 & Mixtral-8x7B & 45.88$\pm$0.71 & 45.12$\pm$0.79 & 43.74$\pm$0.87 & 47.15$\pm$0.96 & 44.30$\pm$1.04 & 47.72$\pm$1.12 \\
 & Llama-3-8B   & 48.23$\pm$1.21 & 45.41$\pm$1.29 & 40.85$\pm$1.37 & 41.00$\pm$1.46 & 44.66$\pm$1.54 & 47.04$\pm$1.62 \\
\midrule
\multirow{4}{*}{TriviaQA}
 & Qwen2.5-7B   & 35.44$\pm$2.67 & 31.71$\pm$1.71 & 39.00$\pm$3.18 & 35.08$\pm$2.10 & 33.79$\pm$2.62 & 33.99$\pm$3.00 \\
 & gpt-oss-20B  & 39.49$\pm$2.09 & 40.85$\pm$2.80 & 44.86$\pm$1.74 & 44.22$\pm$2.26 & 41.23$\pm$3.19 & 39.06$\pm$1.88 \\
 & Mixtral-8x7B & 48.21$\pm$1.62 & 58.42$\pm$1.78 & 59.33$\pm$2.05 & 58.64$\pm$2.33 & 59.21$\pm$2.41 & 56.09$\pm$2.74 \\
 & Llama-3-8B   & 46.11$\pm$1.76 & 49.57$\pm$2.58 & 48.07$\pm$1.92 & 46.28$\pm$2.04 & 45.66$\pm$2.93 & 47.90$\pm$3.10 \\
\bottomrule
\end{tabular}
}
\vspace{-5pt}
\caption{F1 scores across benchmarks for different agents under various prompting strategies. We report mean and standard deviation.}
\label{tab:f1_results}
\vspace{-10pt}
\end{table*}

\begin{table*}[t]
\centering
\resizebox{\textwidth}{!}{
\begin{tabular}{c c cccccc}
\toprule
\textbf{Benchmark} & \textbf{Agent} & \textbf{Raw} & \textbf{CoT} & \textbf{SC} & \textbf{React Reflect} & \textbf{MAD} & \textbf{Summary} \\
\midrule
\multirow{4}{*}{NewsQA} 
 & Qwen2.5-7B   & 40.79$\pm$0.63 & 39.13$\pm$1.34 & 41.91$\pm$0.55 & 41.88$\pm$0.76 & 41.29$\pm$1.57 & 42.08$\pm$0.32 \\
 & gpt-oss-20B  & 37.14$\pm$1.41 & 36.93$\pm$0.39 & 35.22$\pm$1.22 & 36.03$\pm$0.47 & 37.75$\pm$0.83 & 38.17$\pm$1.10 \\
 & Mixtral-8x7B & 28.18$\pm$0.43 & 27.78$\pm$1.02 & 29.27$\pm$0.59 & 29.09$\pm$0.36 & 27.89$\pm$1.31 & 27.24$\pm$1.48 \\
 & Llama-3-8B   & 36.84$\pm$0.74 & 38.12$\pm$0.28 & 40.31$\pm$1.15 & 37.06$\pm$0.91 & 38.72$\pm$0.67 & 37.81$\pm$0.52 \\
\midrule
\multirow{4}{*}{HotpotQA}
 & Qwen2.5-7B   & 46.83$\pm$1.36 & 46.20$\pm$1.42 & 45.80$\pm$1.48 & 43.86$\pm$1.60 & 43.17$\pm$1.54 & 44.14$\pm$1.66 \\
 & gpt-oss-20B  & 50.83$\pm$1.00 & 56.20$\pm$1.06 & 53.80$\pm$1.12 & 54.86$\pm$1.24 & 55.17$\pm$1.18 & 53.14$\pm$1.30 \\
 & Mixtral-8x7B & 44.83$\pm$0.28 & 44.20$\pm$0.34 & 39.80$\pm$0.40 & 44.86$\pm$0.52 & 46.17$\pm$0.46 & 44.14$\pm$0.58 \\
 & Llama-3-8B   & 43.83$\pm$0.64 & 42.20$\pm$0.70 & 44.80$\pm$0.76 & 42.86$\pm$0.88 & 42.17$\pm$0.82 & 41.14$\pm$0.94 \\
\midrule
\multirow{4}{*}{2Wiki}
 & Qwen2.5-7B   & 32.26$\pm$1.30 & 33.73$\pm$1.36 & 31.29$\pm$1.42 & 35.70$\pm$1.48 & 33.31$\pm$1.54 & 34.68$\pm$1.60 \\
 & gpt-oss-20B  & 62.19$\pm$0.94 & 64.80$\pm$1.00 & 61.21$\pm$1.06 & 65.77$\pm$1.12 & 63.24$\pm$1.18 & 63.75$\pm$1.24 \\
 & Mixtral-8x7B & 36.05$\pm$0.23 & 33.94$\pm$0.29 & 34.07$\pm$0.35 & 35.92$\pm$0.41 & 33.09$\pm$0.47 & 36.89$\pm$0.53 \\
 & Llama-3-8B   & 38.12$\pm$0.59 & 36.87$\pm$0.65 & 31.14$\pm$0.71 & 32.85$\pm$0.77 & 37.17$\pm$0.83 & 36.82$\pm$0.88 \\
\midrule
\multirow{4}{*}{TriviaQA}
 & Qwen2.5-7B   & 29.87$\pm$1.41 & 26.16$\pm$0.47 & 33.22$\pm$1.96 & 28.25$\pm$0.85 & 27.92$\pm$1.28 & 25.91$\pm$1.61 \\
 & gpt-oss-20B  & 30.88$\pm$0.92 & 33.19$\pm$1.24 & 36.14$\pm$0.57 & 34.82$\pm$1.09 & 31.71$\pm$1.83 & 31.11$\pm$0.62 \\
 & Mixtral-8x7B & 39.12$\pm$0.44 & 46.88$\pm$0.59 & 49.17$\pm$0.73 & 46.21$\pm$1.03 & 47.09$\pm$0.88 & 42.86$\pm$1.17 \\
 & Llama-3-8B   & 39.92$\pm$0.51 & 42.18$\pm$1.36 & 39.11$\pm$0.66 & 38.14$\pm$0.79 & 37.87$\pm$1.52 & 41.23$\pm$1.68 \\
\bottomrule
\end{tabular}
}
\vspace{-5pt}
\caption{Exact Match (EM) scores across benchmarks for different agents under various prompting strategies. We report mean and standard deviation.}
\label{tab:em_results}
\vspace{-15pt}
\end{table*}

\subsection{Baselines}
\label{appendix:Baselines}

\noindent\textbf{Agent Designs. } 
To capture the breadth of agentic strategies in contemporary LLM research, we incorporate six representative agent designs. \textit{Raw} serves as the direct prompting baseline, providing a view of the backbone’s unaugmented capability. \textit{Chain-of-Thought (CoT)}~\citep{wei2022chain} elicits intermediate reasoning steps, exposing latent deductive processes that enhance multi-step problem solving. \textit{Self-Consistency (SC)}~\citep{wang2023selfconsistency} improves over CoT by sampling multiple reasoning paths and aggregating the most consistent outcome, thereby mitigating variance from any single trajectory. \textit{ReAct-Reflection}~\citep{yao2023react, shinn2023reflexion} augments reasoning with external actions and iterative self-correction, producing more grounded and robust responses. \textit{Multi-Agent Debate (MAD)}~\citep{du2024mad} introduces interactive deliberation among multiple agents, enabling consensus formation through adversarial discussion. Finally, \textit{Multi-Agent Summary} represents collaborative generation where diverse agents contribute partial reasoning that is distilled into a unified answer. These designs span from single-agent prompting to multi-agent coordination, and their diversity is essential for motivating the need for principled routing across heterogeneous strategies.  

\noindent\textbf{Simple Heuristic Ensembling. } 
Based on the agent designs, for baselines we first consider straightforward heuristic ensembling methods. While algorithmically simple, these baselines are crucial because they reveal what can be achieved without sophisticated routing. The \textit{best LLM method} reflects the strongest single backbone performance, serving as a natural lower bound for aggregation strategies, as many prior works focus only on LLM routing, overlooking the potential improvement brought by agent designs. \textit{Average score performance} and \textit{majority vote} represent intuitive ways of pooling outputs across agents---either by averaging confidence scores or relying on democratic consensus. Finally, the \textit{best Agent} baseline always selects the single best-performing agent (per task), simulating an optimal and competitive strategy, this is the upper bound of the performance without agent collaboration with each other. Collectively, these heuristics demonstrate how much performance gain can be extracted from raw diversity alone, providing a necessary contrast against more advanced routing frameworks.

\noindent\textbf{Major Routing Baselines. } 
Beyond heuristics, we benchmark against three state-of-the-art LLM routing frameworks. \textbf{LLM-Blender}~\citep{jiang2023llm} leverages an LLM-as-a-judge paradigm: candidate outputs from multiple agents are presented to a meta-LLM, which adjudicates and selects the final answer. This approach highlights the potential of reflective meta-reasoning but incurs high cost and latency due to repeated LLM calls. \textbf{HybridLLM}~\citep{ding2024hybridllm} adopts a hybrid strategy that combines lightweight scoring heuristics with selective meta-LLM adjudication, aiming to balance efficiency and effectiveness. Finally, \textbf{GraphRouter}~\citep{feng2024graphrouter} formulates routing as a graph-based learning problem, where agents and inputs are embedded into a structured representation, and routing decisions are learned via graph neural networks. This framework demonstrates the benefits of explicitly modeling relational structures among agents and instances, setting a strong precedent for graph-based collaboration learning.

It is worth noting, our work is greatly inspired by the broader idea of graph-based reasoning for LLM orchestration in GraphRouter. However, our work departs fundamentally from GraphRouter in several ways. First, GraphRouter encodes a query and its context as a single node linked to candidate LLMs, effectively reducing the rich internal structure of the context into a coarse representation. In contrast, we decompose queries and contexts into entity-level knowledge graphs, enabling more nuanced modeling of semantic relations that prior research has shown to be critical for reasoning-intensive QA. Second, whereas GraphRouter formulates routing as a naive edge prediction problem between queries and LLMs, our method leverages graph-supervised edge weighting to directly learn collaboration strategies among agents, thereby producing a principled voting scheme rather than a single-model routing decision. Finally, GraphRouter is designed solely for LLM selection, whereas we operate in a multi-agent setting, showing that even when agents share the same backbone, their design choices lead to drastically different behaviors on different tasks. By jointly modeling heterogeneous agents and LLMs, our framework enables emergent collaboration schemes that significantly surpass the strongest single-agent baselines. Also importantly, unlike GraphRouter, our focus is on maximizing accuracy and transferability, rather than cost-effectiveness, offering a complementary perspective in the landscape of graph-based LLM coordination.

\subsection{Training Details}
\label{appendix:Training}
\paragraph{Split.}
We use four datasets—2WikiMultihopQA, HotpotQA, NewsQA, and TriviaQA—with a uniform slicing protocol across splits. For each dataset, we take the \emph{first 500} examples of the official training split as our training data. From the official validation split, we use the \emph{first 100} examples as our validation set and the interval \emph{[100, 200)} (i.e., the next 100 examples) as our test set. This fixed, index-based selection makes experiments easy to replicate across all corpora and avoids leakage between splits while keeping evaluation costs manageable.
\paragraph{Hyperparameters.}
Unless otherwise specified, we train the RouterGNN with hidden size $256$ and $2$ layers, optimized by AdamW (learning rate $1\times10^{-4}$, weight decay $1\times10^{-4}$) and gradient clipping at $1.0$. The device is chosen automatically among \texttt{cuda}/\texttt{mps}/\texttt{cpu} (default \texttt{cpu}). Evaluation uses a stable top-$k$ selection over agents with $k{=}24$ by default (sorting by $(-p,i)$ to deterministically break ties) and weighted voting for answer aggregation. To construct the soft routing targets, we compute per-agent F1 against gold answers and apply $\mathrm{softmax}(\mathrm{F1}/\tau)$ with $\tau{=}0.25$ and a small label smoothing $\varepsilon{=}10^{-3}$. For LLM-backed supervision and logging, we cache agent answers and, when enabled, call the API with temperature $0.2$, max tokens $3000$.
\paragraph{Miscellaneous.}
We adopt an early-stopping practice that saves a checkpoint only when the validation F1 strictly improves; ties do not overwrite the earlier best. All reported test numbers come from the validation-best checkpoint. In addition, during graph construction we enrich question nodes with lightweight type cues: a heuristic keyword scan maps tokens such as \emph{which, where, who, when, why, how, whether} to coarse categories (e.g., person, location, time, reason, manner, yes–no), and when dataset-provided question types are available (e.g., in 2Wiki/HotpotQA) we merge them as well. These type features serve as a weak prior for the model’s type-conditioned routing head without changing supervision targets. Finally, we use Together AI to call the LLM APIs and one NVIDIA GeForce RTX 3090 GPU to train the graph neural networks.

\begin{figure}[t]
	\centering
	\includegraphics[width=1\linewidth]{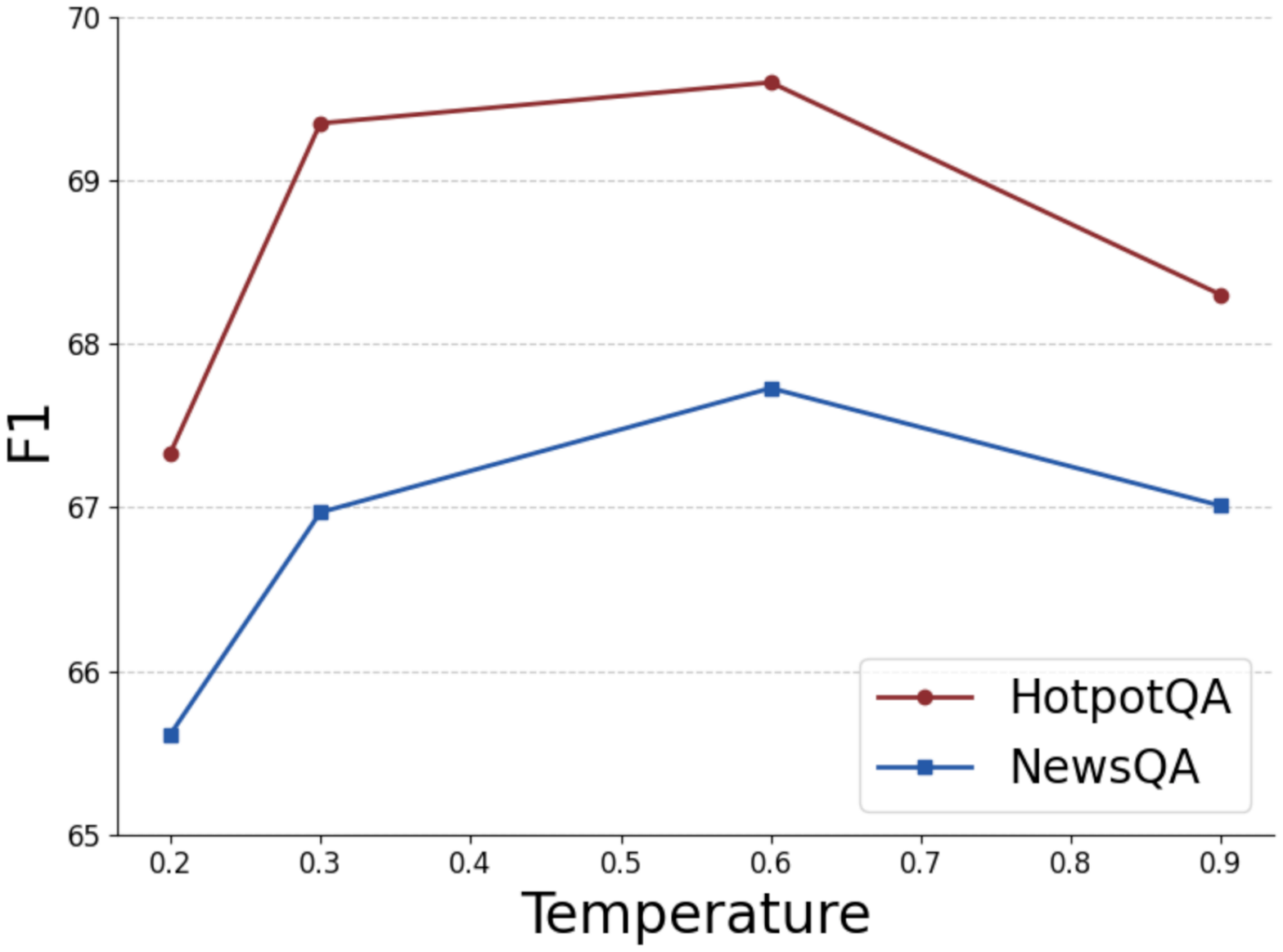}
        \vspace{-20pt}
	\caption{F1 performance on HotpotQA and NewsQA under different temperature settings. Moderate values ($0.3$--$0.6$) yield the strongest results.}
        \vspace{-15pt}
    \label{fig:temp}
\end{figure}

\section{Additional Experiments}
\label{appendix:additional}

In this section, we report the additional experiments we performed to support our claims in the main sections. First, we report the original agent performance used to calculate the average baseline, the best LLM and the best agent, as can be seen in Table~\ref{tab:f1_results} and Table~\ref{tab:em_results}. Then, we also report the effect of other hyper-parameters here. Specifically, we also study the effect of LLM temperature as an example. As shown in Figure~\ref{fig:temp}, we observe that moderate settings ($0.3$--$0.6$) achieve the best balance across different benchmarks, while overly low or high temperatures lead to degraded performance. This indicates that controlled randomness is beneficial for agent diversity, but excessive variance harms consistency.

\section{Prompt Design}

To demonstrate the exact instructions used in our system, we present the full set of prompts that guided the different agent roles. Figures~\ref{fig:qa-prompts-part1} and \ref{fig:qa-prompts-part2} provide a complete overview. These prompts are not intended as a novel design contribution, but rather as transparent documentation of the configurations employed in our experiments.

\begin{figure*}[t]
\centering
\begin{tcolorbox}[colback=myred!5!white,
                  colframe=myred!75!black,
                  title=QA/Reasoning Prompt Suite --- Part I]
{
\textbf{raw\_model:}\\
Given a question, a news context, and retrieved documents, answer the question.\\
The final answer must appear on the last line in the format: \texttt{\textbackslash boxed\{<answer>\}}

\vspace{6pt}
\textbf{cot\_model:}\\
You are a multi-hop reasoning expert and an expert QA agent. Given a question, and the context, think step-by-step. The final answer must appear on the last line in the format: \texttt{\textbackslash boxed\{<answer>\}}

\vspace{6pt}
\textbf{debate\_debater\_a:}\\
You are Debater A. Your goal is to propose the most plausible answer using the provided context.\\
\hspace*{1em}- Make ONE clear claim (the candidate answer).\\
\hspace*{1em}- Support it with 1--2 ultra-short quotes (verbatim substrings) and name the hops.\\
\hspace*{1em}- Explain the link between the quotes in $\leq$2 sentences.\\
Do NOT use outside knowledge and do NOT output the final boxed answer. Make your answer really short and concise.

\vspace{6pt}
\textbf{debate\_debater\_b:}\\
You are Debater B. Your goal is to stress-test A's claim using ONLY the provided context.\\
\hspace*{1em}- If A's quotes or hops are weak, inconsistent, or incomplete, point it out and give corrected quotes/hops.\\
\hspace*{1em}- If a better candidate exists, state ONE alternative with 1--2 short quotes and $\leq$2 sentences of reasoning.\\
\hspace*{1em}- If A is already well-supported, briefly confirm but add one missing check.\\
Do NOT use outside knowledge and do NOT output the final boxed answer. Make your answer really short and concise.

\vspace{6pt}
\textbf{debate\_judge:}\\
You are the Judge. Read A and B as supporting analyses and decide the best final answer using ONLY the given context.\\
If evidence is thin, still make your best context-based guess.\\
Output MUST include nothing but brief final answer in the format: \texttt{\textbackslash boxed\{\}}.

\vspace{6pt}
\textbf{react:}\\
You are a multi-hop reasoning expert and an expert QA agent.\\
Given a question, a news context, and retrieved documents, think step-by-step, silently chain facts to derive a thinking plan,\\
then use this plan to derive the final brief answer.\\
Your output format MUST be a brief final answer on the last line in the format: \texttt{\textbackslash boxed\{<answer>\}}.

\vspace{6pt}
\textbf{reflect:}\\
You are a judge overseeing a multi-hop reasoning expert and an expert QA agent.\\
Given a question, a news context, and retrieved documents, you will evaluate the agent's answer based on the correctness and notes.\\
If the answer is incorrect or incomplete, provide constructive feedback and suggest specific revisions to improve the answer.\\
If the answer is correct and complete, indicate that no further revisions are needed.\\
Your output MUST end with either:\\
\hspace*{1em}- ``Status: revise'' followed by specific feedback and revision suggestions, if the answer needs improvement.\\
\hspace*{1em}- ``Status: final'' if the answer is correct and complete.\\
If you indicate ``Status: revise'', also include a short ``Feedback: <your feedback here>'' section before the final answer.
}
\end{tcolorbox}

\vspace{-10pt}
\caption{Prompt suite for multi-hop QA (Part I).}
\label{fig:qa-prompts-part1}
\end{figure*}

\begin{figure*}[t]
\centering
\begin{tcolorbox}[colback=myred!5!white,
                  colframe=myred!75!black,
                  title=QA/Reasoning Prompt Suite --- Part II]
{\large
\textbf{think\_a:}\\
You are a multi-hop reasoning expert and an expert QA agent.\\
Given a question, a news context, and retrieved documents, think step-by-step, chain facts to derive the answer.\\
Give your final answer as a single entity, and a concise reasoning process that leads to the answer.

\vspace{6pt}
\textbf{think\_b:}\\
You are a multi-hop reasoning expert and an expert QA agent.\\
Given a question, a news context, and retrieved documents, think step-by-step, chain facts to derive the answer.\\
Give your final answer as a single entity, and a concise reasoning process that leads to the answer.

\vspace{6pt}
\textbf{summarize:}\\
You are the multi-hop reasoning expert and an expert QA agent. You receive outputs from other agents. Use them as \textbf{supporting signals}.\\
If A and B agree on the same short span, return it. If they differ, pick the best answer with your own reasoning.\\
Your output format MUST end with the brief final answer on the last line in the format: \texttt{\textbackslash boxed\{<answer>\}}.

\vspace{6pt}
\textbf{router:}\\
You are a routing model.\\
Decide if the following user question is CHALLENGING or EASY.\\
Answer with a single token: \texttt{'CHALLENGING'} or \texttt{'EASY'}.

\vspace{6pt}
\textbf{note:}\\
Here are the rules you must \textbf{STRICTLY} follow:\\
1. Always return the answer as the SHORTEST exact entity only. The answer is always within 10 words, and usually within 5 words.\\
2. If the question is yes/no, respond strictly with \emph{yes} or \emph{no} only.\\
3. For year ranges, never use hyphens; instead, use ``from XXXX to YYYY'' or ``XXXX until YYYY''.\\
4. Do not output sentences, explanations, or phrases with verbs; the answer must be a single entity expression only.\\
5. One way or another, you must return your best guess, and the final answer must be in the format: \texttt{\textbackslash boxed\{<answer>\}}.
}
\end{tcolorbox}

\vspace{-10pt}
\caption{Prompt suite for multi-hop QA (Part II).}
\label{fig:qa-prompts-part2}
\end{figure*}

\section{Case Study}

To better understand the effectiveness of our routing mechanism, we conducted a set of qualitative case studies such as those shown in Figure~\ref{fig:agent-routing-three-examples}. Before discussing the results, it is useful to categorize the typical types of agent errors we observe. These include: (1) \emph{content errors}, where the answer text is semantically incorrect; (2) \emph{format errors}, where the output does not follow the required answer format (e.g., missing a boxed final answer or producing extraneous text); (3) \emph{mixed errors}, where the response contains partially correct information but is polluted with unrelated or contradictory details; and (4) \emph{null responses}, where the agent fails to provide an answer at all (e.g., returning ``None'' or empty output). These diverse error modes highlight the importance of an intelligent router that can identify which agents are more likely to generate useful and valid responses for a given question.

The three representative examples illustrate how our model’s routing module successfully shifts probability mass toward the most reliable agents. 

In the graph construction process, we first parse the context into a heterogeneous graph of entities and their dependency or semantic relations. Then, for each \textbf{backbone $\times$ agent}, we run an \textbf{LLM-as-judge} process: given the question, the local entity subgraph, and the agent’s prompt or capability profile, the judge scores each (agent, entity) pair and establishes \texttt{manage} edges for the highest-scoring pairs — meaning ``this agent is responsible for these entities.''

\begin{figure}[t]
\centering
\definecolor{entityblue}{RGB}{66,133,244}
\definecolor{agentorange}{RGB}{245,124,0}
\definecolor{questiongreen}{RGB}{46,125,50}
\definecolor{relgray}{RGB}{150,150,150}
\definecolor{managec}{RGB}{0,151,167}
\definecolor{qrefc}{RGB}{213,0,0}
\definecolor{q2agentc}{RGB}{123,31,162}

\begin{tikzpicture}[
  xshift=-1cm,
  >=Latex,
  font=\scriptsize,
  node distance=6mm and 8mm,
  entity/.style={circle, draw=entityblue, thick, minimum size=5mm, inner sep=0pt, fill=entityblue!5},
  agent/.style={circle, draw=agentorange, thick, minimum size=6mm, inner sep=0pt, fill=agentorange!10},
  qnode/.style={circle, draw=questiongreen, thick, minimum size=7mm, inner sep=0pt, fill=questiongreen!10},
  rel/.style={draw=relgray, dashed, opacity=0.6, ->},
  manage/.style={draw=managec, very thick, ->},
  qref/.style={draw=qrefc, very thick, ->},
  q2agent/.style={draw=q2agentc, thick, ->}
]

\node[qnode] (Q1) at (-4, 5.2) {};
\node[above=1pt of Q1] {Q1: Who is known as `the Falcon'?};

\node[agent] (A1) at (-9.3, 3.6) {};
\node[agent] (A2) at (-5.9, 3.8) {};
\node[agent] (A3) at (-1.9, 3.6) {};
\node[agent] (A4) at ( 2.3, 3.4) {};

\node[above=1pt of A1]  {gpt\_oss\_20b::raw};
\node[right=1pt of A2]  {qwen2p5\_7b::cot};
\node[right=1pt of A3]  {llama3\_8b::summary};
\node[above=1pt of A4]  {mixtral\_8x7b::react};

\node[entity] (E11)  at (-9.9,  0.9)  {E11};  
\node[entity] (E2)  at (-8.9, -1.3)  {E2};  
\node[entity] (E3)  at (-6.7,  0.1)  {E3};  
\node[entity] (E6)  at (-4.9,  0.8)  {E6};  
\node[entity] (E5)  at (-4.1, -0.9)  {E5};  
\node[entity] (E8)  at (-2.1,  1.0)  {E8};  
\node[entity] (E7)  at (-1.3, -1.4)  {E7};  
\node[entity] (E4)  at ( 0.5, -0.2)  {E4};  
\node[entity] (E9)  at ( 3.7,  1.1)  {E9};  
\node[entity] (E10) at ( 3.1, -1.0)  {E10}; 
\node[entity] (E1) at (-6.1,  2.0)  {E1}; 
\node[entity] (E12) at (-0.1,  2.1)  {E12}; 

\node[below=1pt of E1]  {The Falcon Takes Over};
\node[below=1pt of E2]  {The Falcon Steps Out};
\node[below=1pt of E3]  {Irving Reis};
\node[below=1pt of E4]  {The Gay Falcon};
\node[below=1pt of E5]  {A Date with the Falcon};
\node[below=1pt of E6]  {George Sanders};
\node[below=1pt of E7]  {Gay Lawrence};
\node[left=1pt of E8]  {\textbf{the Falcon}};
\node[below=1pt of E9]  {1942};
\node[below=1pt of E10] {1941};
\node[below=1pt of E11] {black-and-white mystery film};
\node[below=1pt of E12] {B film};

\draw[rel] (E1) -- (E2);
\draw[rel] (E1) -- (E11);
\draw[rel] (E11) -- (E3);
\draw[rel] (E1) -- (E6);
\draw[rel] (E6) -- (E7);
\draw[rel] (E7) -- (E8);
\draw[rel] (E12) -- (E4);
\draw[rel] (E12) -- (E5);
\draw[rel] (E4) -- (E10);
\draw[rel] (E5) -- (E10);

\draw[qref]    (Q1) -- (E8);
\draw[q2agent] (Q1) -- (A1);
\draw[q2agent] (Q1) -- (A2);
\draw[q2agent] (Q1) -- (A3);
\draw[q2agent] (Q1) -- (A4);

\draw[manage] (A1) -- (E1);
\draw[manage] (A1) -- (E2);

\draw[manage] (A2) -- (E1);
\draw[manage] (A2) -- (E11);

\draw[manage] (A3) -- (E4);
\draw[manage] (A3) -- (E8);
\draw[manage] (A3) -- (E6);
\draw[manage] (A3) -- (E1);

\draw[manage] (A4) -- (E12);
\draw[manage] (A4) -- (E9);
\draw[manage] (A4) -- (E10);

\node (LEG) [anchor=south west, draw, rounded corners=2pt, inner sep=3pt, align=left, fill=white]
  at ([xshift=5mm,yshift=8mm]Q1.north west) {%
  \textbf{Legend}\\[2pt]
  \textbf{1}:\ \tikz{\draw[rel] (0,0)--(8mm,0);} \ Entity--Entity dependency (gray dashed)\\[4pt]
  \textbf{2}:\ \tikz{\draw[manage] (0,0)--(8mm,0);} \ \texttt{manage} (agent$\rightarrow$entity, teal)\\[4pt]
  \textbf{3}:\ \tikz{\draw[q2agent] (0,0)--(8mm,0);} \ \texttt{q2agent} (question$\rightarrow$agent, purple)\\[4pt]
  \textbf{4}:\ \tikz{\draw[qref] (0,0)--(8mm,0);} \ \texttt{q\_ref} (question$\rightarrow$entity, red)
};

\node (CTX) [anchor=south west, draw, rounded corners=2pt, inner sep=3pt, align=left, fill=white, text width=5.2cm]
  at ([xshift=0mm,yshift=5mm]LEG.north west) {%
  \textbf{Context}\\[2pt]
  \textit{Title: The Falcon Takes Over.} The Falcon Takes Over (also known as "The Falcon Steps Out") is a 1942 black-and-white mystery film directed by Irving Reis. The B film was the third, following "The Gay Falcon" and "A Date with the Falcon" (1941), to star George Sanders as the character Gay Lawrence, a gentleman detective known by the sobriquet "the Falcon".\\
  \textbf{Gold}\\[2pt]
  Gay Lawrence.
};

\node[anchor=south east, draw, rounded corners=2pt, inner sep=3pt, align=left, fill=white, text width=5.2cm]
  at ([xshift=-6mm,yshift=8mm]Q1.north west) {%
  \textbf{Entity--Entity Relations (Drawn Only)}\\[4pt]
  \textbf{E1 → E2} dep:alias \\
  \textbf{E1 → E11} dep:attr \\
  \textbf{E11 → E9} nummod \\
  \textbf{E1 → E6} star \\
  \textbf{E6 → E7} prep:as \\
  \textbf{E12 → E4} prep:following \\
  \textbf{E7 → E8} dep:attr \\
  \textbf{E12 → E5} prep:following \\
  \textbf{E4 → E10} appos \\
  \textbf{E5 → E10} appos
};

\end{tikzpicture}

\caption{An illustrative example of the heterogeneous graph structure used in our framework. 
The graph consists of three types of nodes: \textbf{question nodes} (green, top), representing input questions; 
\textbf{agent nodes} (orange, middle), corresponding to different reasoning or retrieval agents; 
and \textbf{entity nodes} (blue, bottom), representing extracted knowledge items such as people, films, time expressions, or attributes. 
Edges capture various types of relations between these nodes: dashed gray edges denote linguistic or semantic relations between entities, 
teal edges indicate \texttt{manage} links connecting agents to the entities they are responsible for, 
purple edges represent \texttt{q2agent} connections from the question to potentially useful agents, 
and red edges (\texttt{q\_ref}) link questions directly to referenced entities.}
\label{fig:spread-entities}
\end{figure}
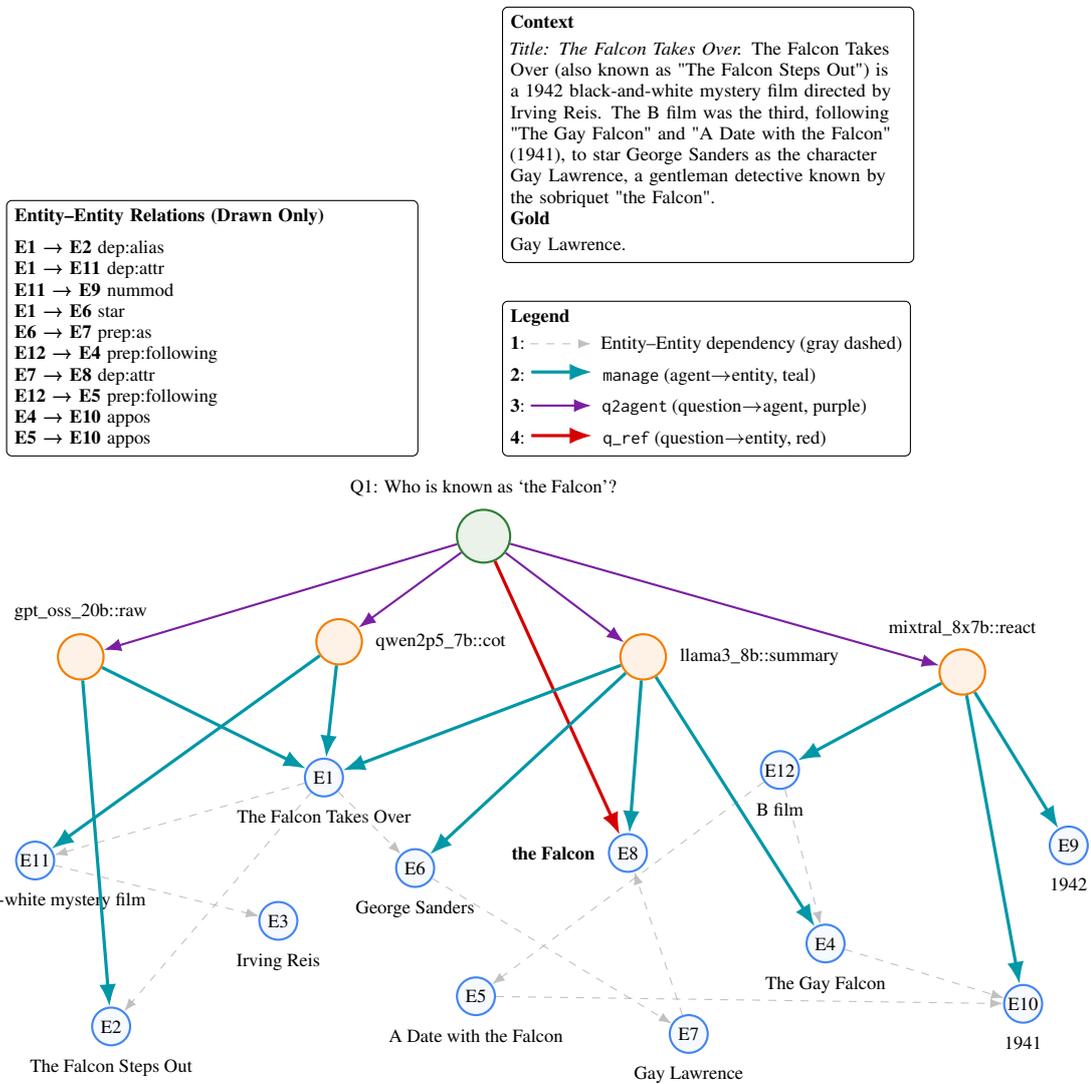

In this example (Figure~\ref{fig:spread-entities}), the judge connects llama3\_8b::summary to key entities like \textit{The Falcon Takes Over}, and \textit{George Sanders}. The selected agent can then reason step by step through its managed subgraph:

\textit{The Falcon Takes Over} $\xrightarrow{\texttt{star}}$ \textit{George Sanders} $\xrightarrow{\texttt{prep:as}}$ \textit{Gay Lawrence} $\xrightarrow{\texttt{dep:attr}}$ \textit{the Falcon}.

By following this reasoning chain grounded in the context, the agent is able to reach the correct answer node, therefore receiving the message.

\begin{figure*}[t]
\centering
\begin{tcolorbox}[colback=myblue!5!white,
                  colframe=myblue!75!black,
                  title=Agent Routing Examples (Three QA Cases)]
{

\textbf{Example 1}\\
\textbf{Question:} What use British singer Gary?\\
\textbf{Gold Answer:} iPhone apps\\
\textbf{Model Prediction:} iPhone apps

\vspace{4pt}
\textbf{Top-4 Agents}\\
\texttt{BACKBONE::qwen2p5\_7b\_turbo::AGENT::raw} $\rightarrow$ \emph{iPhone apps} (0.062086940)\\
\texttt{BACKBONE::gpt\_oss\_20b::AGENT::mad} $\rightarrow$ \emph{iPhone apps} (0.060009051)\\
\texttt{BACKBONE::gpt\_oss\_20b::AGENT::raw} $\rightarrow$ \emph{iPhone apps} (0.059776004)\\
\texttt{BACKBONE::gpt\_oss\_20b::AGENT::react\_reflect} $\rightarrow$ \emph{iPhone apps} (0.059702747)\\[4pt]
\textbf{Last-4 Agents}\\
\texttt{BACKBONE::mixtral\_8x7b::AGENT::mad} $\rightarrow$ \emph{iPhone, keyboard, Twitter} (0.003830930)\\
\texttt{BACKBONE::mixtral\_8x7b::AGENT::raw} $\rightarrow$ \emph{iPhone/keyboard} (0.004338742)\\
\texttt{BACKBONE::mixtral\_8x7b::AGENT::sc} $\rightarrow$ \emph{iPhone, keyboard, microphone} (0.004404212)\\
\texttt{BACKBONE::mixtral\_8x7b::AGENT::cot} $\rightarrow$ \emph{iPhone/music} (0.004915467)

\vspace{10pt}
\textbf{Example 2}\\
\textbf{Question:} What caused the accident?\\
\textbf{Gold Answer:} the bus lost control on a curve\\
\textbf{Model Prediction:} the bus lost control on a curve

\vspace{4pt}
\textbf{Top-4 Agents}\\
\texttt{BACKBONE::qwen2p5\_7b\_turbo::AGENT::cot} $\rightarrow$ \emph{the bus lost control on a curve} (0.081045948)\\
\texttt{BACKBONE::qwen2p5\_7b\_turbo::AGENT::react\_reflect} $\rightarrow$ \emph{the bus lost control on a curve} (0.079336852)\\
\texttt{BACKBONE::qwen2p5\_7b\_turbo::AGENT::raw} $\rightarrow$ \emph{the bus lost control on a curve} (0.078211986)\\
\texttt{BACKBONE::qwen2p5\_7b\_turbo::AGENT::mad} $\rightarrow$ \emph{the bus lost control on a curve} (0.071492545)\\[4pt]
\textbf{Last-4 Agents}\\
\texttt{BACKBONE::mixtral\_8x7b::AGENT::mad} $\rightarrow$ \emph{Bus left highway, rolled over} (0.003636589)\\
\texttt{BACKBONE::mixtral\_8x7b::AGENT::summary} $\rightarrow$ \emph{Bus left highway, rolled over} (0.005194495)\\
\texttt{BACKBONE::mixtral\_8x7b::AGENT::sc} $\rightarrow$ \emph{Lost control on curve} (0.005307447)\\
\texttt{BACKBONE::mixtral\_8x7b::AGENT::raw} $\rightarrow$ \emph{unknown} (0.005436183)

\vspace{10pt}
\textbf{Example 3}\\
\textbf{Question:} Which film has the director who died later, The Fatal Mistake or The Devil's Hairpin?\\
\textbf{Gold Answer:} The Devil's Hairpin\\
\textbf{Model Prediction:} The Devil's Hairpin

\vspace{4pt}
\textbf{Top-4 Agents}\\
\texttt{BACKBONE::gpt\_oss\_20b::AGENT::cot} $\rightarrow$ \emph{The Devil's Hairpin} (0.124811694)\\
\texttt{BACKBONE::gpt\_oss\_20b::AGENT::sc} $\rightarrow$ \emph{The Devil's Hairpin} (0.124024361)\\
\texttt{BACKBONE::gpt\_oss\_20b::AGENT::raw} $\rightarrow$ \emph{The Devil's Hairpin} (0.123276740)\\
\texttt{BACKBONE::gpt\_oss\_20b::AGENT::react\_reflect} $\rightarrow$ \emph{The Devil's Hairpin} (0.120928459)\\[4pt]
\textbf{Last-4 Agents}\\
\texttt{BACKBONE::mixtral\_8x7b::AGENT::mad} $\rightarrow$ \emph{The Fatal Mistake} (0.008660691)\\
\texttt{BACKBONE::llama3\_8b\_lite::AGENT::mad} $\rightarrow$ \emph{The Fatal Mistake} (0.009608842)\\
\texttt{BACKBONE::mixtral\_8x7b::AGENT::cot} $\rightarrow$ \emph{The Devil's Hairpin} (0.009781577)\\
\texttt{BACKBONE::mixtral\_8x7b::AGENT::sc} $\rightarrow$ \emph{The Fatal Mistake} (0.009870600)
}
\end{tcolorbox}

\vspace{-8pt}
\caption{Per-question agent routing snapshots for three examples. Each case shows the question, gold answer, model prediction, and the four highest- and lowest-probability agents with their generated answers.}
\label{fig:agent-routing-three-examples}
\end{figure*}

\onecolumn

\end{document}